\documentclass[authoryear, 11pt]{elsarticle}
\journal{Transportation Research Part C: Emerging Technologies}
\usepackage[sc]{mathpazo}
\usepackage{geometry}
\usepackage{color}
\usepackage{array}
\usepackage{bbding}
\usepackage[fleqn]{amsmath}
\usepackage{amssymb}
\usepackage{amsthm}
\usepackage{graphicx}
\usepackage[ruled]{algorithm2e}
\usepackage{algpseudocode}  
\usepackage{natbib}
\usepackage{lscape}
\usepackage[colorlinks,
linkcolor=blue, 
anchorcolor=blue,
citecolor=blue,
]{hyperref}

\usepackage{comment}
\usepackage{bm}
\usepackage[title]{appendix}
\usepackage{longtable}
\usepackage{mathtools}
\usepackage{chngcntr}

\usepackage{enumerate}
\usepackage{cuted}
\usepackage{multirow}
\usepackage{makecell}
\usepackage{booktabs}
\usepackage{ragged2e}
\usepackage{url}
\usepackage{tabularx}
\makeatletter
\allowdisplaybreaks

\@ifundefined{showcaptionsetup}{}{%
 \PassOptionsToPackage{caption=false}{subfig}}
\usepackage{subfig}
\makeatother

\usepackage{babel}
\usepackage{enumitem}
\usepackage{enumerate}

\begin{document}\sloppy

\begin{frontmatter}

\title{LIFT: Interpretable truck driving risk prediction with literature-informed fine-tuned LLMs}

\author[tsinghua]{Xiao Hu}
\author[tsinghua]{Yuansheng Lian}
\author[ppsuc]{Ke Zhang}
\author[BJUT]{Yunxuan Li}
\author[SKLab]{Yuelong Su}
\author[tsinghua]{Meng Li\corref{corLi}}

\cortext[corLi]{Corresponding author.}
\ead{mengli@tsinghua.edu.cn}

\address[tsinghua]{Department of Civil Engineering, Tsinghua University, Beijing, China}
\address[ppsuc]{College of Traffic Management, People's Public Security University of China, Beijing, China}
\address[BJUT]{Beijing Key Laboratory of Traffic Engineering, Beijing University of Technology, Beijing, China}
\address[SKLab]{State Key Laboratory of Intelligent Green Vehicle and Mobility, Tsinghua University, Beijing, China}

\begin{abstract}
\noindent 
This study proposes an interpretable prediction framework with literature-informed fine-tuned (LIFT) LLMs for truck driving risk prediction. The framework integrates an LLM-driven \textit{Inference Core} that predicts and explains truck driving risk, a \textit{Literature Processing Pipeline} that filters and summarizes domain-specific literature into a literature knowledge base, and a \textit{Result Evaluator} that evaluates the prediction performance as well as the interpretability of the LIFT LLM. After fine-tuning on a real-world truck driving risk dataset, the LIFT LLM achieved accurate risk prediction, outperforming benchmark models by 26.7\% in recall and 10.1\% in F1-score. Furthermore, guided by the literature knowledge base automatically constructed from 299 domain papers, the LIFT LLM produced variable importance ranking consistent with that derived from the benchmark model, while demonstrating robustness in interpretation results to various data sampling conditions. The LIFT LLM also identified potential risky scenarios by detecting key combination of variables in truck driving risk, which were verified by PERMANOVA tests. Finally, we demonstrated the contribution of the literature knowledge base and the fine-tuning process in the interpretability of the LIFT LLM, and discussed the potential of the LIFT LLM in data-driven knowledge discovery.\par

\hfill\break
\end{abstract}

\end{frontmatter}

\section{Introduction}
\label{s:Introduction}
\noindent
Truck driving risk is of great significance to traffic safety. While trucks play an important role in economic development, truck-involved accidents tend to have high fatality rates and cause substantial material damages \citep{zhou2019analysis, niu2020risk, baikejuli2022study, zainuddin2023influence}. In current industry practices, logistics companies monitor risky driving events (e.g., forward collision warnings) through onboard devices and manually monitor drivers when such events occur frequently, followed by safety interventions via phone calls or other methods \citep{zhang2024driving}. However, as truck drivers often work for extended periods in isolation, real-time and effective supervision across all their working scenarios is exceedingly difficult. Meanwhile, truck driving risk is often under the influence of multi-factors, traditional frequency-based risk identification approaches fail to reveal the underlying causes of the risky events, making it difficult for intervention personnel to promptly assess drivers' conditions and deliver appropriate countermeasures \citep{baikejuli2022study}. \par

To uncover the complex interplay of factors underlying truck driving risk and achieve precise prediction, researchers have developed extensive statistical analytic methods \citep{behnood2019timeofday, hu2025influencing}, machine learning (ML) methods \citep{xue2023contextaware, wang2024commercial}, and integrated methods combining the advantage of statistical analysis and ML approaches \citep{yuan2022using, jin2023realtime}. However, these data-driven methods are fundamentally sensitive to data distribution. As it has been found that truck drivers may exhibit different driving behavior across different traffic scenarios \citep{hu2025influencing}, when the data collection conditions change, the prediction and interpretation results obtained from the traditional data-driven models may vary significantly. Yet, collecting data from as many types of traffic environments as possible is often cost-prohibitive.\par

To overcome data limitations, scholars have begun leveraging pre-trained large language models (LLMs) to address complex prediction problems in the domain of traffic prediction \citep{zhang2025large}. As LLMs have shown strong generalization capabilities in predictive tasks among different domains, it has been found that a fine-tuning process with a small amount of domain-specific data is sufficient to enable a general-purpose LLM to demonstrate strong performance in traffic prediction problems \citep{li2024urbangpt}. Furthermore, different with traditional black-box prediction methods, LLMs can infer the underlying information behind the data points using semantic reasoning, which enhances the interpretability of their prediction results. For example, LLMs may identify a driver's fatigue state based on repeated instances of risky driving behavior. However, general-purpose LLMs may show hallucinations in understanding specific domain problems. Thus, researchers have integrated human-constructed domain knowledge of the research problem to inform the LLM with domain concepts \citep{guo2024explainable, liu2025collision, xie2025could}. However, in actual truck operation scenarios, the influencing factors of truck driving risks demonstrate high complexity and diversity in different traffic scenarios, which can hardly be formulated in a standardized domain knowledge base constructed by human experts.\par
 
\begin{figure}[htb]
\centering
\includegraphics[width=\textwidth]{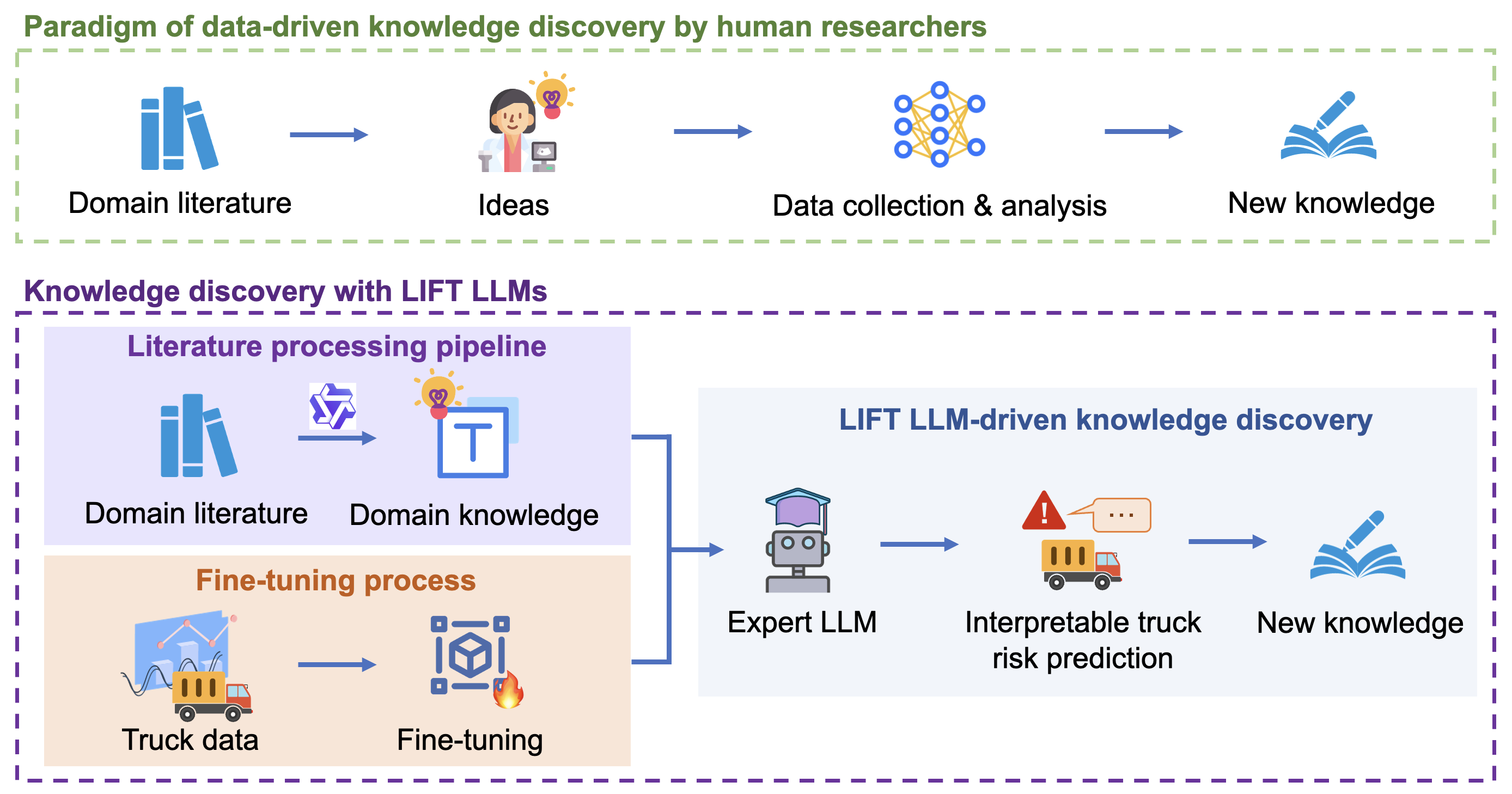}
\caption{Knowledge discovery process with LIFT LLMs compared with the paradigm of data-driven knowledge discovery by human researchers.}
\label{fig:LIFT intro}
\end{figure}

Just as human researchers acquire knowledge from previous literature (Figure \ref{fig:LIFT intro}), LLMs can also learn domain-specific knowledge from a vast amount of specialized literature to enhance their capabilities in corresponding research areas. In the field of truck safety research, researchers have accumulated a substantial body of literature over the past decades, which includes extensive qualitative and quantitative conclusions derived from data analysis in various traffic scenarios. Thus, the previous literature containing information that goes beyond what can be captured in expert-defined knowledge bases. As LLMs show great capability in understanding long-context semantic data, it becomes possible that LLMs can directly learn from the literature and enhance its capability in dealing with domain task, such as generating research ideas \citep{guo2025automating}. However, few research has been done to explore the possibility of augmenting LLMs' capability in interpretable prediction tasks with domain literature.\par

Thus, in this research, we propose a novel interpretable prediction framework with literature-informed fine-tuned (LIFT) LLMs. As shown in Figure \ref{fig:LIFT intro}, the framework fine-tunes an LLM using real-world truck driving data to realize precise truck driving risk prediction. Furthermore, unlike existing studies that manually defines domain knowledge bases by human experts, we design an LLM-driven literature processing pipeline that automatically construct a domain knowledge base from hundreds of research papers, which enhances the capability of the fine-tuned LLM in result interpretation. With the domain knowledge base, the fine-tuned LLM can interpret truck driving risk and discover new knowledge such as risky scenarios.\par

Specifically, we constructed a truck driving risk dataset aggregating real-time driving behavior data and environmental information such as real-time traffic speed, utilizing real-world truck trajectory data and risky events detected by on-board devices. Using well-designed prompt template, we converted the truck driving risk dataset into a text-based dataset and fine-tuned an open-source LLM. The prediction results of the fine-tuned LLM outperformed benchmark methods. Meanwhile, given the domain literature knowledge base, the literature-informed fine-tuned LLM was employed to explain the driving risk by pointing out the key variables and combination of variables that contribute to the high driving risk in the high risk trajectory samples. Feature importance was calculated for each influencing factor by calculating the frequency of that feature detected by the LLM as key variable in the dataset. Results show that the ranking result of the feature importance aligns with that derived from the widely-adopted random forest model. Furthermore, the LIFT LLM identified complex variable combinations that may contribute to high truck driving risk, which are verified through PERMANOVA tests, highlighting the potential of the LIFT LLM in discovering risky traffic scenarios. Finally, we illustrated the inherent robustness of the LIFT LLM in result explanation with the domain literature knowledge base, comparing with traditional data-driven methods. \par

The main contributions of this paper are as follows:\par

\begin{itemize}
    \item Firstly, we propose a novel interpretable prediction framework with literature-informed fine-tuned (LIFT) LLMs for truck driving risk prediction. This framework integrates a literature processing pipeline which automatically turns hundreds of domain papers into a literature knowledge base, and fine-tunes an LLM to achieve high capability in truck driving risk prediction and interpretation. \par

    \item Secondly, utilizing a real-world truck driving risk dataset, we validated the prediction performance of the LIFT LLM. Results show that the LIFT LLM outperforms benchmark machine learning models by 26.7\% in recall and over 10\% in F1-score.\par

    \item Thirdly, through prompting the LIFT LLM to identify key variables and key variable combinations for high risk trajectory samples, we demonstrated its capability in result interpretation and risk scenario discovery. And the results were verified by benchmark comparison and PERMANOVA tests for examining variable differences between risky and non-risky samples.\par 

\end{itemize}

The rest of the paper is organized as follows. Section \ref{s:Literature Review} reviews previous research on truck driving risk prediction and application of LLMs in previous traffic safety research. Section \ref{s:Methodology} introduces the design of the proposed interpretable prediction framework with LIFT LLM. Section \ref{s:Experimental settings} discusses the dataset preparation and experimental settings. Section \ref{s:Result analysis} discusses the results and analysis. Section \ref{s:Discussion} provides further discussion on benefit of LIFT LLM in the industry and its potential in knowledge discovery. Finally, Section \ref{s:Conclusions} concludes the paper and discusses the limitation of this research and direction for future research.\par

\section{Literature Review}
\label{s:Literature Review}
\noindent

\subsection{Truck driving risk prediction}
\label{ss:Truck driving risk prediction}

To enhance truck driving safety, researchers have developed various models to predict driving risk in different scenarios \citep{shao2023adaptive, guo2021study, ma2023dynamic}. Among different truck driving risks, forward collision risk represents a particularly critical concern due to the substantial mass of trucks and the consequent severe damage potential in collision events. Existing research has conducted extensive studies on short-term forward collision risk prediction \citep{tan2022risk, shangguan2023empirical}. For example, real-time prediction models of forward collision risks in car-following behaviors have been developed based on onboard video \citep{xie2024personalized, wang2022effect}. Based on the prediction results, these studies support onboard warning systems to inform truck drivers to keep safety following distances \citep{shao2023adaptive}.\par

However, short-term risk warnings based on car-following distance metrics (including Time-To-Collision, TTC) provide limited reaction time for drivers and suffers from the inherent trade-off between recall rate and false alarm rate. That is, to ensure the effectiveness of warning systems by identifying more true (risky) cases, these systems often employ conservative warning strategies. This approach, however, leads to frequent warning triggers, which not only distract drivers but also erode their trust in the warning system \citep{bao2023evaluation}. To address these limitations, researchers have developed more sophisticated approaches by incorporating richer data sources to predict driving risk on at the trip scale. Specifically, these approaches leverage comprehensive information about driver behavior patterns, driving styles, and context information such as real-time traffic conditions to predict trip-scale driving risks in specific scenarios, such as on a certain road link, thereby enabling proactive warning systems that provide both early alerts and recommendations on driving strategy and route choices \citep{arbabzadeh2018datadriven, ma2023prediction, masello2023using}.

For example, \citet{mehdizadeh2021predicting} integrated five dynamic data sources (driver characteristics, weather, time periods, etc.) with ensemble learning to predict 30-minute ahead high-risk events on a certain truck trip. And \citet{yu2024riskformer} predicted the real-time crash risk of a vehicle using the sequential risk behaviors happened during the trip. Due to the diverse data requirements and challenges in data collection, driving risk prediction at the trip scale has not yet been sufficiently studied. Accurately predicting driving risk under the combined influence of multiple factors from multiple data sources remains a significant challenge. To bridge the gap, this study focuses on truck driving risk prediction at the trip scale, utilizing multi-source data collected from long-term operation of a real-world truck fleet.\par

\subsection{Interpretable prediction models for truck driving risk}
\label{ss:Interpretable prediction models for truck driving risk}
\noindent

Traditional statistical methods can effectively reveal the impact effects and significance of different factors on driving risk. However, constrained by generalized linear modeling framework, traditional statistical methods struggle to achieve accurate risk prediction at the trip scale, particularly when handling complex multi-source feature inputs and extensive nonlinear relationships \citep{mehdizadeh2021predicting}. To overcome these challenges, machine learning-based approaches have been developed. Early work employed support vector machines and decision tree-based ensemble learning methods \citep{savelonas2020classification}. More recently, deep learning architectures have been widely adopted in traffic risk prediction \citep{moosavi2021driving, jin2023realtime, gao2025stcmgcn}.

Machine learning's powerful fitting capabilities enable modeling more sophisticated risk factors than human design regression models, effectively capturing complex dynamic and static driving behavior characteristics \citep{moosavi2023contextaware, zhang2024highrisk}. Besides, machine learning can also predict driving risk of a driver or crash risk in a certain spatial-temporal region by effectively capturing high-dimensional temporal patterns without manual feature engineering \citep{ma2023realtime, zhang2023critical, han2024transformerbased}. Yet, as machine learning models rely on latent high-dimensional features, which cannot be intuitively understood by human, this black-box nature creates interpretability challenges, undermining trust among stakeholders. Existing solutions include post-hoc interpretability methods such as feature importance analysis of tree-based methods \citep{mehdizadeh2021predicting, xia2024investigating} and SHAP (SHapley Additive exPlanations) for deep learning models \citep{ma2023prediction, jin2023realtime, masello2023using}. 

However, these post hoc interpretation methods fundamentally differ from statistical model interpretations of variable effects and significance. Post hoc approaches require computation based on the training data distribution, demonstrating how different variable values influence final predictions \citep{lundberg2017unified}. Consequently, the interpretative results are inherently sensitive to data distributions—particularly when heterogeneity exists within the data, where sampling bias may lead to inconsistent explanations across different subpopulations. Furthermore, while post hoc methods prove valuable for aggregate-level risk factor analysis, they cannot provide instance-specific intuitive explanations during real-time prediction. This critical limitation hinders their practical application for explaining predictions in real-time risk forecasting scenarios.\par

\subsection{Large language model for interpretable risk prediction}
\label{ss:Large language model for interpretable risk prediction}
\noindent

In recent years, the rapid advancement of Large Language Models (LLMs) has attracted growing attention to their applications in traffic prediction problems \citep{zheng2023chatgpt, mahmud2025integrating}. As LLMs are pre-trained on massive amounts of corpus data, they demonstrate strong generalization capabilities across problems with different data distributions. Meanwhile, compared to traditional machine learning models, LLMs exhibit unique advantages in flexibly processing diverse complex inputs and generating intuitive natural language responses in a real-time manner, thereby demonstrating significant potential for multi-source data-driven traffic risk prediction. Researchers have developed LLM-based interpretable prediction methods that demonstrate remarkable performance and flexibility across various traffic prediction tasks \citep{zhang2025large, peng2025lcllm}. Since LLMs typically use tokens as their basic input and output units, applying them to traffic risk prediction requires converting traffic scenario information into token representations. Current research has developed two primary conversion approaches for interpretable prediction: (1) encoding qualitative descriptions of traffic scenes into predefined tokens, and (2) representing semantic information of traffic scenarios through natural text encoding.\par

The first approach converts continuous variables into discrete categories (e.g., high and low) using manually defined thresholds and domain knowledge, assigning unique tokens to each categorical value. This transforms traffic scenarios into token sequences, allowing risk prediction to be framed as a text classification task. For instance, \citet{xie2025could} employed fuzzy logic to discretize autonomous vehicle dynamics and surrounding traffic relationships into qualitative tokens, then trained a BERT model to classify collision risks from these token sequences. However, this method requires manual rule design and problem-specific retraining, limiting its flexibility.\par

Another approach encodes traffic scenarios as textual descriptions, leveraging pre-trained LLMs' semantic processing capabilities. Researchers use open-source LLMs with their native tokenizers to convert scenario text into tokens, framing risk prediction as text generation problem. For instance, \citet{liu2025collision} transformed vehicle dynamics and traffic features into text via prompt templates, then fine-tuned Llama-3-8B-Instruct using LoRA to generate risk predictions and explanations. They designed a reasoning pipeline where the LLM interprets quantitative variables' semantic meanings (e.g., "executing a sharp left turn") before generating risk predictions and explanations for given traffic scenarios. However, the generated explanations primarily rely on the LLM's semantic understanding capability pretrained from massive text data from the Internet, which may lack specialized knowledge in traffic risk assessment and potentially lead to hallucinated interpretations. \citet{guo2024explainable} mitigated this problem by enriching prompts with expert-defined traffic rules and variable definitions to guide interpretations. Though flexible, such text-based approaches still require manual input of knowledge and remain vulnerable to hallucinations.\par

Beyond these two approaches, some studies have employed pretrained LLMs as base model while training additional encoders specifically for traffic scenario data. These encoders directly map input data to the latent space of the LLM's hidden layers to enhance prediction performance and generalization \citep{li2024urbangpt}. However, this approach relies on black-box encoders, making the overall model just as inexplicable as traditional machine learning methods. Consequently, it cannot enable semantic-based interpretability as direct LLM applications \citep{liu2025collision, guo2024explainable}.\par

\subsection{Research gap}
\label{ss:Research gap}
\noindent

In summary, current studies have developed various machine learning and LLM-based approaches to capture high-dimensional features in data for accurate truck driving risk prediction under multi-source complex factors. However, existing methods exhibit two key limitations in model interpretability. Firstly, post hoc explanation techniques in conventional machine learning models compute feature importance based on data distributions of the training data, which may yield inconsistent interpretations across different traffic scenarios when heterogeneity exists. Secondly, existing LLM-based risk prediction methods rely on manual input of domain knowledge and rules to enhance model interpretability, resulting in limitations in generalizability and flexibility. Additionally, while prior research has validated LLMs' capabilities in microscopic driving risk prediction such as collision risk in seconds, their potential for trip-scale driving risk prediction remains unexplored.\par

To bridge the gaps, this study proposes an interpretable prediction framework with LIFT LLMs for trip-scale truck driving risk prediction. The proposed framework utilizes textualized real-world data for fine-tuning and prediction to avoid manual rule design on tokenization. Furthermore, the framework integrates a literature processing pipeline to automatically construct a domain literature knowledge base to inform the fine-tuned LLM with domain specific knowledge. Based on the fine-tuning process and the literature knowledge base, the LIFT LLM can not only achieve superior prediction performance, but also output stable explanation on prediction results regardless of data distribution. Furthermore, the LIFT LLM can identify new risky scenarios and explain the truck driving risk in natural language, thereby enhancing the usability of the system for safety interventions in the industry.\par

\section{Methodology}
\label{s:Methodology}
\noindent

\subsection{Framework design}
\label{ss:Framework design}
\begin{figure}[htb]
\centering
\includegraphics[width=\textwidth]{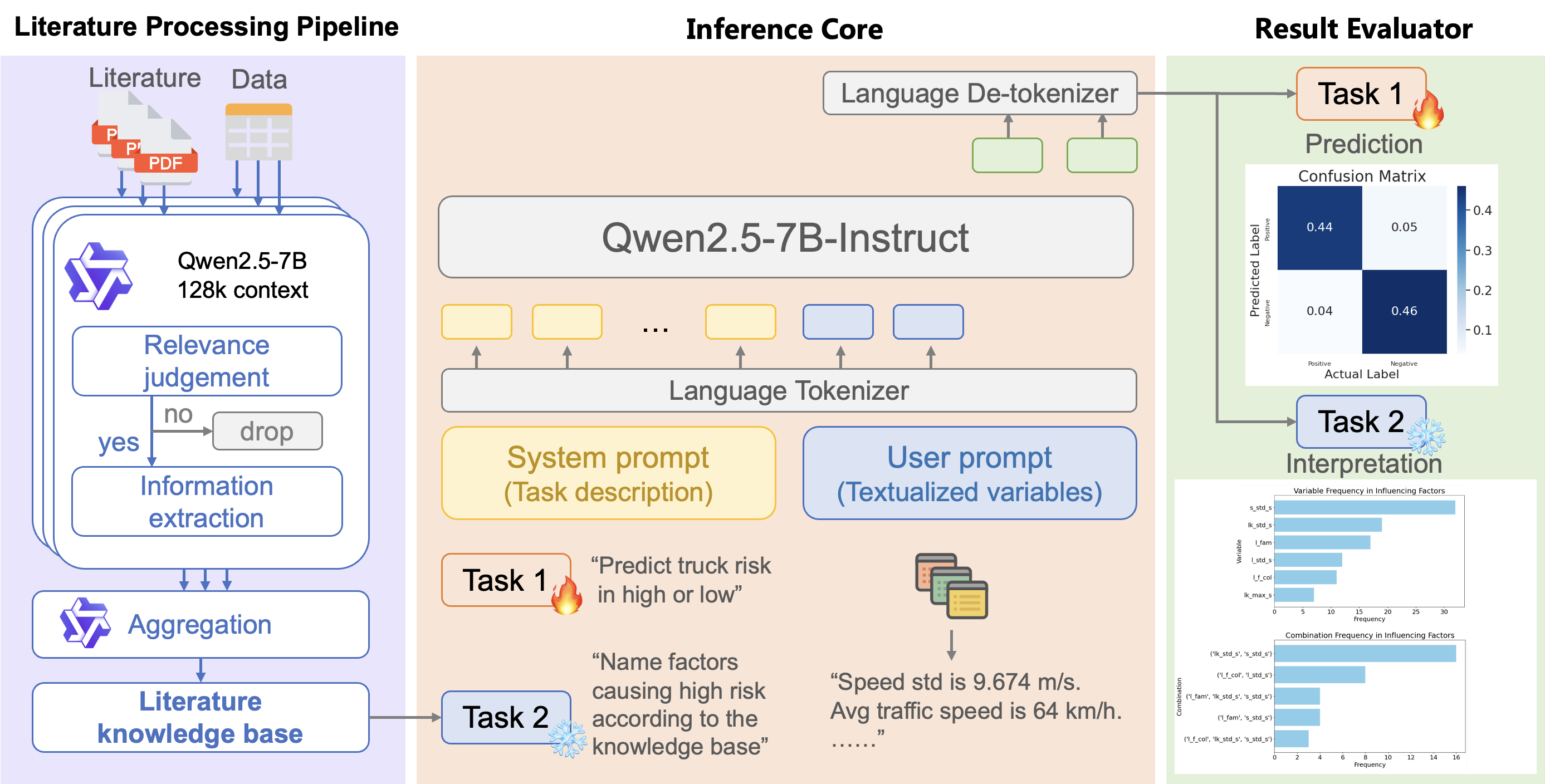}
\caption{The proposed interpretable prediction framework with LIFT LLMs. The framework consists of three components: \textit{Literature Processing Pipeline}, which constructs a literature knowledge base from a large volume of academic papers; \textit{Inference Core}, which performs prediction and explanation of truck driving risk based on a fine-tuned LLM, informed by the literature knowledge base; and \textit{Result Evaluator}, which assesses the model's predictive performance and the quality of its interpretative outputs.}
\label{fig:LIFT framework}
\end{figure}

\noindent
The proposed interpretable prediction framework with LIFT LLM is shown in Figure \ref{fig:LIFT framework}. The framework consists of three components. In the center is the \textit{Inference Core}, which performs prediction and explanation of truck driving risk based on a fine-tuned LLM. As the previous stage of the \textit{Inference Core}, the \textit{Literature Processing Pipeline} extracts relevant research conclusions from domain literature and constructs the literature knowledge base for the \textit{Inference Core}. Finally, the \textit{Result Evaluator} collects the output of the \textit{Inference Core} and assesses the model's predictive performance and the quality of its interpretative outputs.\par

The core of the framework is an LLM, which is fine-tuned using a textualized dataset derived from real-world truck driving risk data. To enhance the LLM's ability to generate interpretable explanations for its predictions, we design the \textit{Literature Processing Pipeline} for constructing a domain literature knowledge base, which considers the relevance between the previous studies with the collected truck driving risk data. The constructed knowledge base is incorporated into the model's prompts to strengthen its understanding of domain-specific knowledge, forming the literature-informed fine-tuned (LIFT) LLM. Finally, we evaluate its performance in prediction and result interpretation on a test dataset in the \textit{Result Evaluator}. \par

Specifically, we decompose the interpretable prediction problem into two sub-tasks. Task 1 is a classification task that involves predicting the level (high or low) of truck driving risk based on given input data. Task 2 is designed to evaluate the interpretability of the predictions: given a high-risk trajectory sample, the model is required to identify the key variables contributing to the high risk. To ensure the desired outputs for both tasks, we design distinct system prompts in the LLM's input for the two tasks. The following sections provide a detailed description of the design of the three modules.\par

\subsection{Inference Core}
\label{ss:Inference Core}
\noindent
In the \textit{Inference Core}, we frame the truck driving risk prediction task as a language modeling problem. By constructing a textualized dataset and applying supervised fine-tuning to a LLM, we enable the model to perform truck driving risk prediction.\par

\subsubsection{Problem description}
\label{sss:Problem description}
\noindent
In our previous research, it is found that truck driving risk is affected by long-term behavior pattern of the truck driver, short-term driving behavior during the trip of the truck driver and real-time traffic environment \citep{hu2025influencing}. Thus, following previous research, we formalize truck driving risk prediction problem as a binary classification prediction task:
\begin{equation}
    P(Risky) = f_\theta(\mathbf{X}_l, \mathbf{X}_s, \mathbf{X}_t),
\end{equation}

where $P(Risky)$ represents the probability of forward collision conflicts occurring on target road segment, $\mathbf{X}_l$ represents the long-term behavior pattern of the truck, $\mathbf{X}_s$ represents the short-term driving behavior of the truck during the trip, $\mathbf{X}_t$ represents the real-time traffic conditions, $f_\theta(\cdot)$ represents the prediction function that maps input features to the interval [0,1], and $\theta$ is the model parameters.

In this work, we employ a LLM as the prediction model $f_\theta(\cdot)$, and a language tokenizer $\mathcal{T}(\cdot)$ is utilized to convert the input features into a sequence of tokens:
\begin{equation}
    \mathbf{t} = \mathcal{T}(\mathbf{X}_l, \mathbf{X}_s, \mathbf{X}_t, \mathbf{K}),
\end{equation}

where $\mathbf{t}$ represents the tokenized sequence of input features, $\mathcal{T}(\cdot)$ is the language tokenizer, and $\mathbf{K}$ is the domain knowledge base which contains the definitions of the input variables, domain knowledge extracted from the literature, and prompt design related information that determines how to convert these variables into token sequences.

Thus, the prediction problem is transformed into a language modeling problem:
\begin{equation}
    P_\theta(t_i|\mathbf{t}_{<i}, \mathbf{K}) = \text{softmax}(h_i/\tau),
    \label{eq:NTP with temperature}
\end{equation}

where $t_i$ represents the $i$-th token in the sequence, $\mathbf{t}_{<i}$ represents all tokens before position $i$, $h_i$ represents the hidden states of the LLM at position $i$, $\tau$ is the temperature parameter for the LLM, and $\mathbf{K}$ represents the domain knowledge base.

The final prediction is obtained by:
\begin{equation}
    \hat{y} = g(\hat{t}),
\end{equation}

where $g(\cdot)$ is a mapping function that converts the generated tokens $\hat{t}$ to the risk probability $\hat{y}$.

\subsubsection{Prompt construction}
\label{sss:Prompt construction}
\noindent
In this study, we incorporate domain knowledge $\mathbf{K}$ directly into the LLM input rather than having human researchers establish prediction models $f(\cdot)$ based on expert knowledge. Specifically, we convert input variables and domain knowledge into prompts that comprehensively include task descriptions, domain knowledge, values and definitions of input variables, and specify the format requirements for model responses. The following Figure \ref{fig:Prompt template} illustrates the prompt template and the expected answer.\par

\begin{figure}[htb]
\centering
\includegraphics[width=\textwidth]{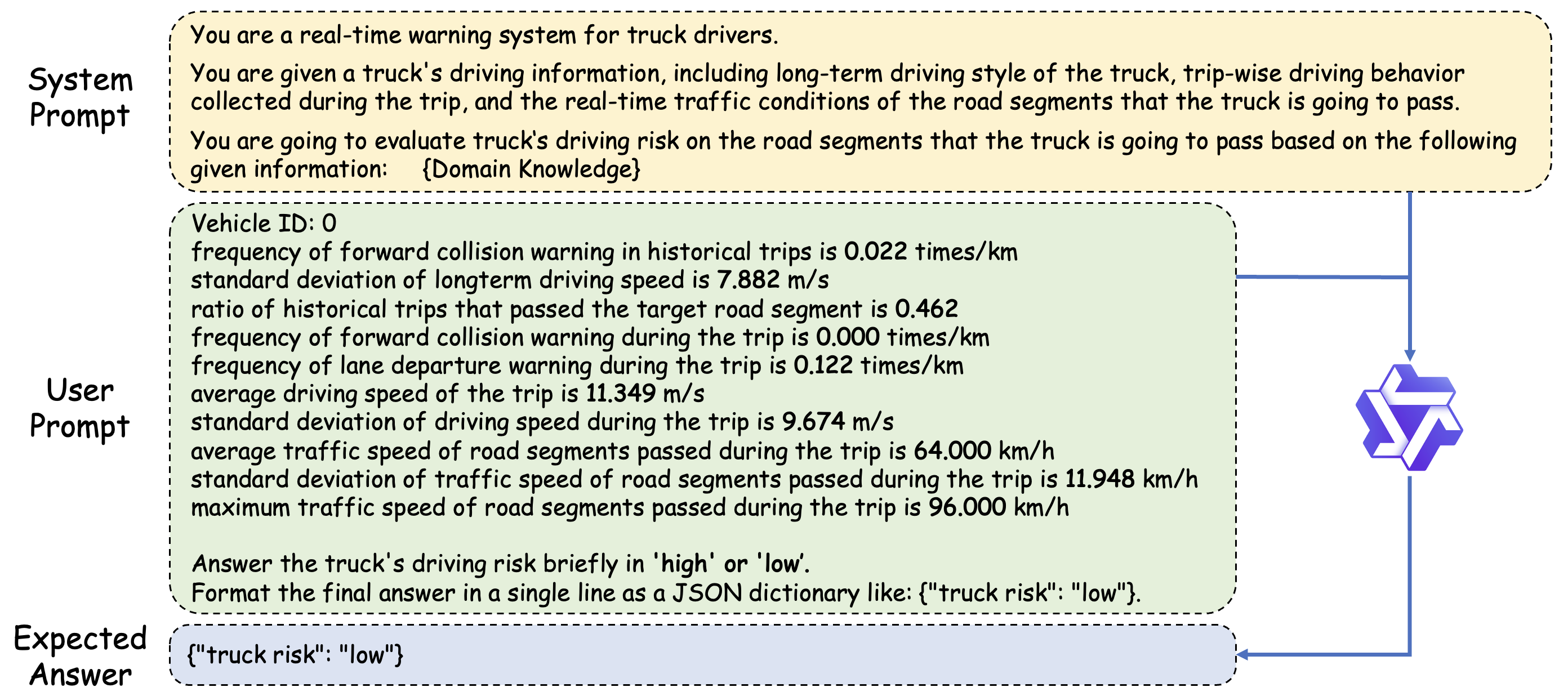}
\caption{Prompt template for truck driving risk prediction problem (Task 1).}
\label{fig:Prompt template}
\end{figure}

As shown in Figure \ref{fig:Prompt template}, the prompt is structured in a hierarchical manner to optimize the model's inference process. System prompts, containing task descriptions and domain knowledge, are positioned at the beginning of the prompt to establish the foundational context and is consistent for each data sample. This is followed by user prompts that contain sample-specific information and format requirements that standardize model output. This hierarchical arrangement leverages the caching mechanism of transformer-based LLMs, optimizing inference efficiency by fixing identical prefixes. Finally, the format requirements ensure that the model can provide outputs that meet the output specifications, supporting model testing and further applications.

\subsubsection{Supervised fine-tuning}
\label{sss:Supervised fine-tuning}
\noindent
In this research, supervised fine-tuning is adopted to adapt an general-purpose large language model to risk prediction task while maintaining its fundamental language capabilities. We firstly convert the original truck driving risk dataset into a text dataset $\mathcal{D}$ using the prompt template shown in Figure \ref{fig:Prompt template} for supervised fine-tuning. The model then learns to predict truck driving risk by optimizing the following objective function for the entire text dataset $\mathcal{D}$:
\begin{equation}
    L^{FT}(\mathcal{D}) = \sum_{d\in\mathcal{D}} \sum_{i=1} logP(t_i|\mathbf{t}_{<i}, \mathbf{K}),
\end{equation}

where $d$ represents a sample in the dataset $\mathcal{D}$, $t_i$ represents the $i$-th token in the sequence, $\mathbf{t}_{<i}$ represents all tokens before position $i$, $\mathbf{K}$ represents the domain knowledge base.

We employ LoRA \citep{hu2021lora} for efficient model adaptation during supervised fine-tuning. LoRA introduces low-rank updates to model parameters, expressing the adapted weight matrix as $W = W_0 + \alpha\Delta W$, where $W_0$ denotes pre-trained weights, $\alpha$ is a scaling factor, and $\Delta W$ represents fine-tuning updates. Rather than directly updating $\Delta W$, LoRA approximates it through the product of two lower-dimensional matrices: $\Delta W = BA$, where $B \in \mathbb{R}^{d\times r}$ and $A \in \mathbb{R}^{r\times k}$. The rank $r$ is chosen to be significantly smaller than $W_0$'s dimensions ($r \ll \min(d,k)$), reducing parameter count from $d\times k$ to $r\times(d+k)$. This technique effectively reduces training resource consumption and helps to avoid over-fitting.\par

\subsection{Literature Processing Pipeline}
\label{ss:Literature Processing Pipeline}
\begin{figure}[htb]
\centering
\includegraphics[width=\textwidth]{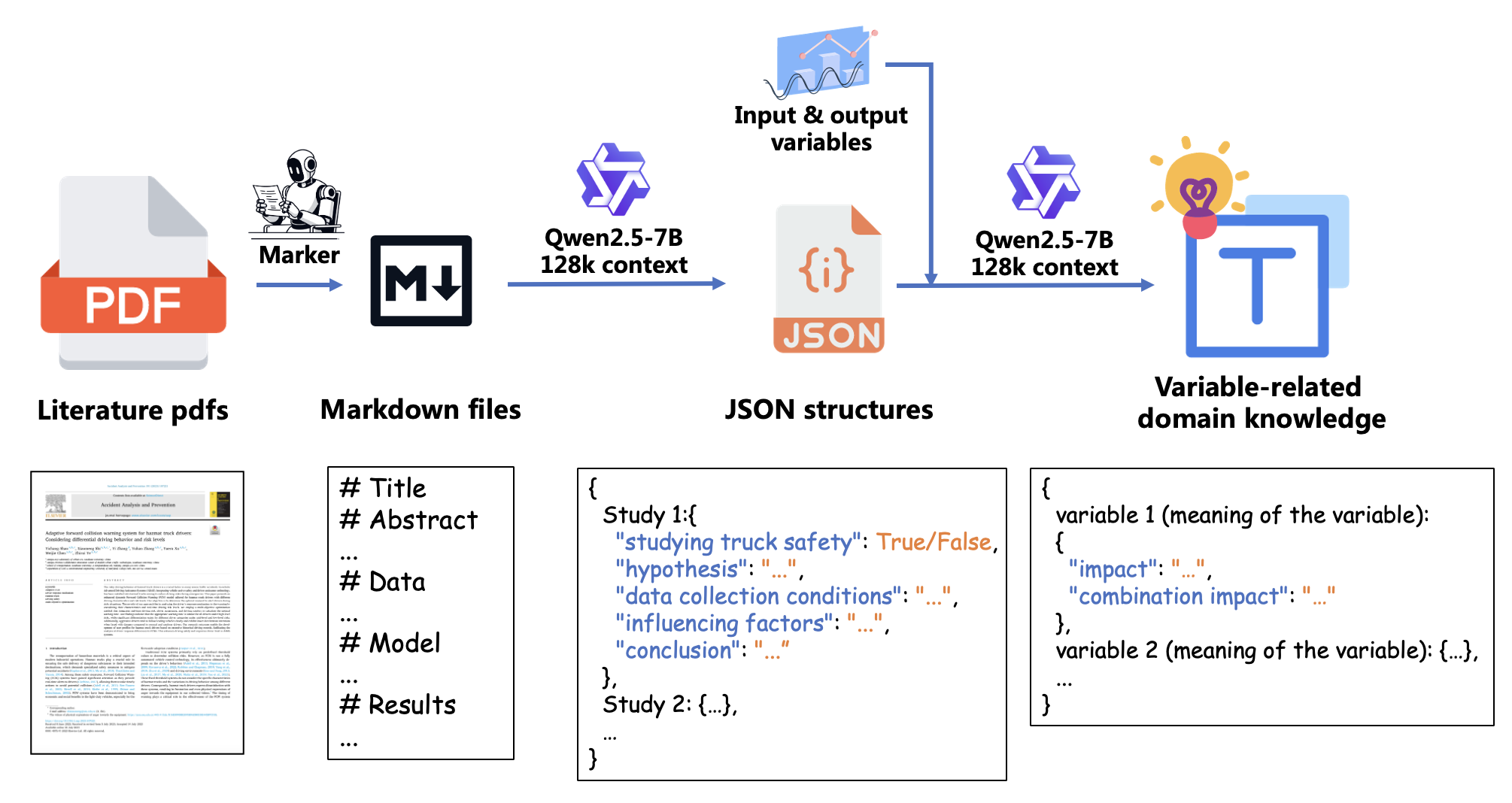}
\caption{\textit{Literature Processing Pipeline} for variable-related domain knowledge extraction.}
\label{fig:Process flow for variable-related domain knowledge extraction}
\end{figure}

\noindent
In this study, we extracted domain knowledge related to input and output variables of the prediction problem from domain literature following the process pipeline as shown in Figure \ref{fig:Process flow for variable-related domain knowledge extraction}. Firstly, we manually searched and downloaded domain research papers in PDF format, and converted the PDF files to Markdown files using an open-source tool marker \citep{paruchuri2025marker}. Converting PDF files into Markdown is necessary because LLMs can only process plain text. Moreover, using the Markdown format offers the following advantages: (1) Markdown is widely used in the pretraining data of LLMs, resulting in strong native support for this format; (2) Markdown preserves structural elements such as headings, tables, lists, and bold text, which facilitates the effective extraction of key information by the LLM.\par

\begin{figure}[htb]
\centering
\includegraphics[width=\textwidth]{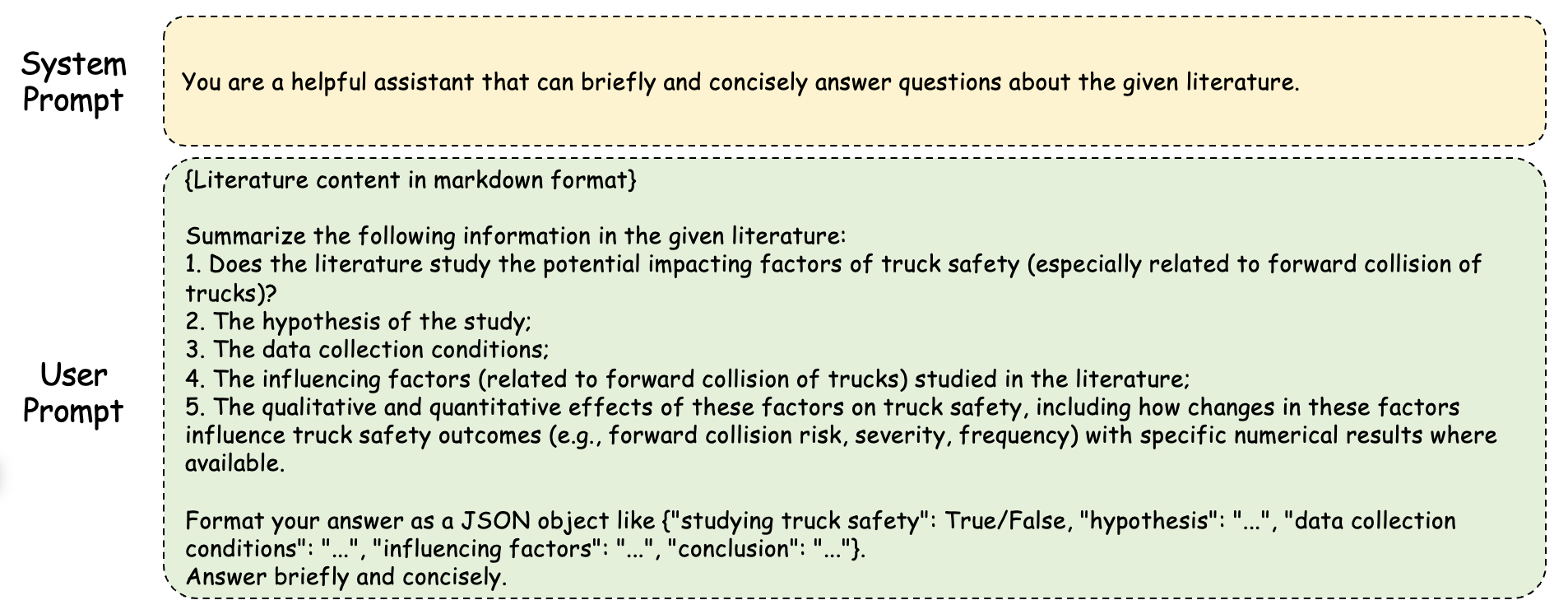}
\caption{Prompt for literature information extraction.}
\label{fig:Prompt for literature information extraction}
\end{figure}

Subsequently, we conducted a screening and information extraction process on the literature using the prompt template shown in Figure \ref{fig:Prompt for literature information extraction}. We used a long-context LLM \citep{qwen2.5} to read each paper in full and determine whether it investigated potential contributing factors to truck safety, particularly forward collision risks, and filtered out irrelevant studies. We then extracted key information from the selected papers, including research hypothesis, data collection conditions, factors analyzed, and analytical conclusions, and saved the results in JSON format.

\begin{figure}[htb]
\centering
\includegraphics[width=\textwidth]{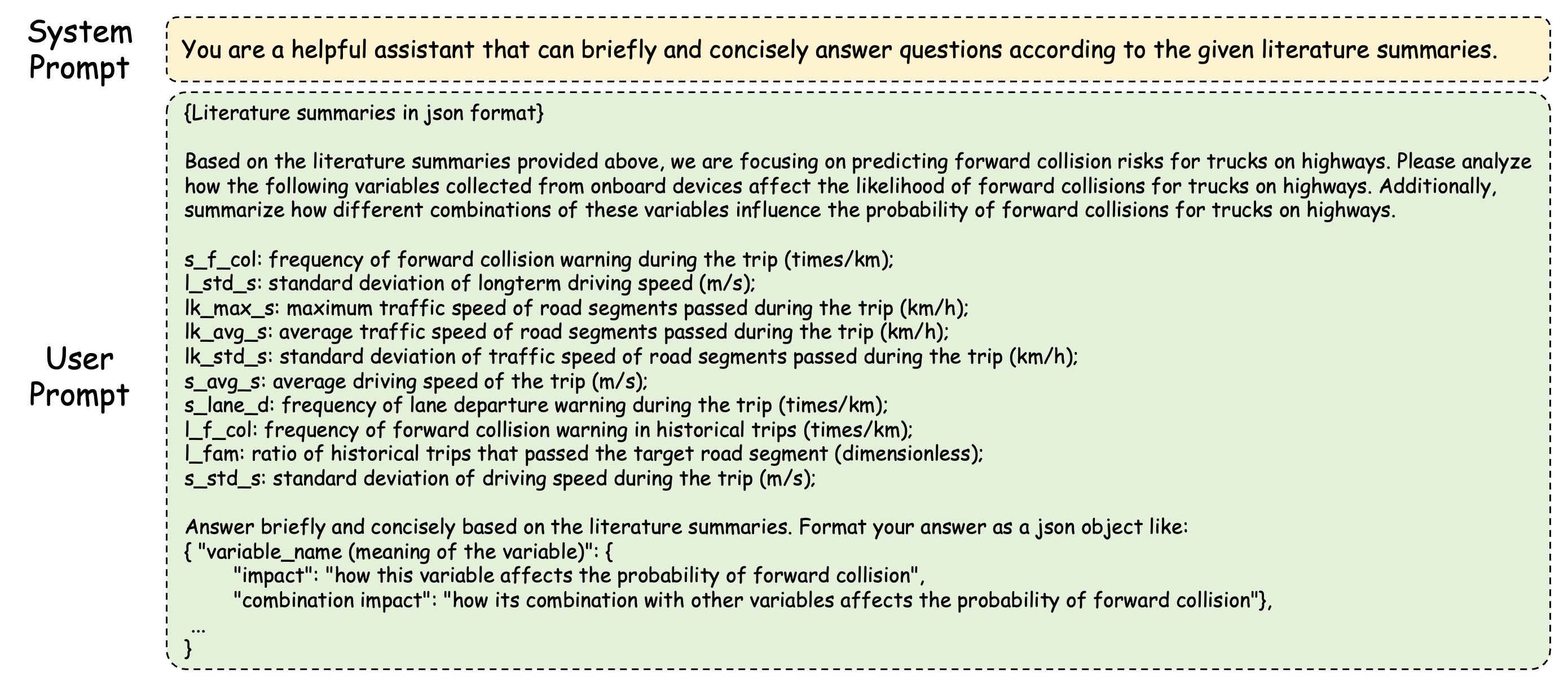}
\caption{Prompt for domain knowledge extraction.}
\label{fig:Prompt for domain knowledge extraction}
\end{figure}

Finally, using the prompt template shown in Figure \ref{fig:Prompt for domain knowledge extraction}, we aggregated the extracted JSON structures, along with definitions of input and output variables involved in this problem, and used a long-context LLM to summarize research conclusions from these literatures related to these input and output variables, saving as a JSON structure. Specifically, considering that combinations of multiple influencing factors may jointly affect truck driving risks, we required the LLM to summarize both qualitative and quantitative impacts of each input variable on driving risks, while also providing how each influencing factor's combination with other factors affects driving risks. And the final extracted JSON structure can be used as the domain knowledge base $\mathbf{K}$ for model input.\par

Using this pipeline, we ensure that the extracted domain knowledge base $\mathbf{K}$ can fully reflect the characteristics of the truck driving risk prediction problem in this study, thus improving the performance of the model in result explanation. \par

\subsection{Result Evaluator}
\label{ss:Result Evaluator}
\noindent
In the interpretable framework, we need to evaluate the performance of the LIFT LLM across two tasks. As introduced in Section \ref{ss:Framework design}, Task 1 involves binary classification of truck driving risks, while Task 2 focuses on evaluating the interpretability of the predictions. \par

\begin{figure}[htb]
\centering
\includegraphics[width=\textwidth]{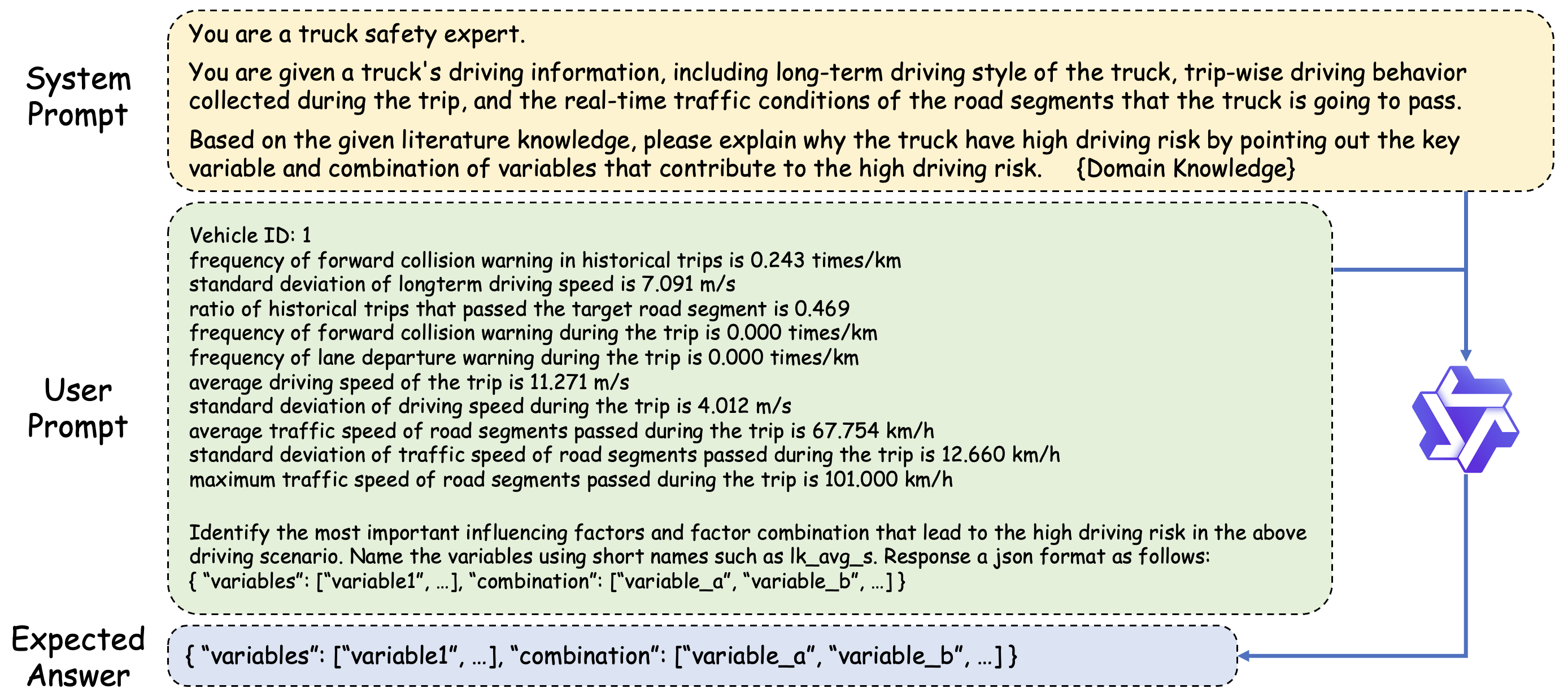}
\caption{Prompt template for key risk factor identification problem (Task 2).}
\label{fig:Prompt template task2}
\end{figure}

For Task 1, we partition the data into training and testing sets, and use the prompt as shown in Figure \ref{fig:Prompt template} to get the prediction result for each test sample. Subsequently, we employ standard evaluation metrics for binary classification problems to assess the model's prediction performance. For Task 2, inspired by existing research on interpretable prediction methods, we evaluate model interpretability through variable importance analysis. Specifically, we leverage the instruction-following capability of the LLM by feeding it high-risk truck trajectory samples and prompting the fine-tuned LLM to output the key variables contributing to the high risk, as shown in Figure \ref{fig:Prompt template task2}. Subsequently, we aggregate the model's identification results of key variables across all high-risk samples in the test set, treating the frequency of variables being identified as key variable as their importance measure to establish a variable importance distribution.\par

\section{Experimental settings}
\label{s:Experimental settings}

\subsection{Data description}
\label{ss:Data description}
\noindent
In this study, we constructed a truck driving risk dataset through multi-source data aggregation. We collected extensive long-term trajectory data from trucks, along with forward collision warning events detected by onboard devices, and combined them with real-time traffic condition data provided by Amap to build the truck driving risk dataset. The data collection was conducted in Dongguan, China, from Jan 4th, 2024 to March 1st, 2024, involving trajectory data from 4,672 trucks. In this dataset,  following our previous work \citep{hu2025influencing}, we used forward collision warning events detected on the target segments of Guanglong Expressway to evaluate driving risk. Specifically, we recorded risk labels as binary variables (0-1) based on whether forward collision warning events occurred on the target segments on Guanglong Expressway for each trajectory. And we used trucks' long-term behavior patterns, short-term driving behavior during the trip, and real-time traffic environment information as input features for the prediction model.\par

To capture the long-term behavior patterns of trucks, we selected trucks that had traversed the target segments at least 60 times during the data collection period, totaling 68 vehicles, and constructed the dataset based on these trajectories. Firstly, for each selected truck, we calculated a series of long-term behavior pattern variables using all historical trajectories. Next, we treated each trajectory of the selected trucks as a data sample, computing short-term driving behavior variables exhibited by trucks from departure until reaching the target segments on Guanglong Expressway. Finally, we collected real-time traffic environment variables on the target segments at the time when the truck passed through, serving as real-time traffic environment variables.

Ultimately, we extracted 1792 trajectories, including 74 trajectories with forward collision warnings on the target segments (risk label = 1) and 1717 trajectories without forward collision warnings on the target segments (risk label = 0). To ensure model simplicity, we selected the 10 influencing factors with the greatest impact on truck driving risk as input features, according to elasticity analysis in our previous work \citep{hu2025influencing}. The distribution of the input features and their abbreviations for each trajectory is shown in Table \ref{tab: Statistical description}.

\begin{table}[!ht]
    \scriptsize
    \centering
    \caption{Statistical description of variables}
    \begin{tabular}{lrrrr}
    \hline
        \textbf{Variable} & \textbf{Mean} & \textbf{Std} & \textbf{Max} & \textbf{Min} \\ \hline
        \textbf{l\_f\_col} (Frequency of forward collision warning in historical trips, times/km) & 0.13  & 0.08  & 0.27  & 0.00   \\ 
        \textbf{l\_std\_s} (Standard deviation of longterm driving speed, m/s) & 7.17  & 0.59  & 8.63  & 6.12   \\ 
        \textbf{l\_fam} (Ratio of historical trips that passed the target road segment ) & 0.38  & 0.09  & 0.49  & 0.08   \\ 
        \textbf{s\_f\_col} (Frequency of forward collision warning during the trip, times/km) & 0.15  & 0.18  & 0.96  & 0.00   \\ 
        \textbf{s\_lane\_d} (Frequency of lane departure warning during the trip, times/km) & 0.05  & 0.08  & 0.58  & 0.00   \\ 
        \textbf{s\_avg\_s} (Average driving speed of the trip, m/s) & 11.55  & 2.61  & 21.54  & 3.19   \\ 
        \textbf{s\_std\_s} (Standard deviation of driving speed during the trip, m/s) & 6.52  & 1.36  & 11.97  & 0.78   \\ 
        \textbf{lk\_avg\_s} (Average traffic speed of road segments passed during the trip, km/h) & 69.09  & 7.20  & 95.56  & 40.83   \\ 
        \textbf{lk\_std\_s} (Standard deviation of traffic speed of road segments passed during the trip, km/h) & 12.63  & 2.94  & 33.52  & 3.23   \\ 
        \textbf{lk\_max\_s} (Maximum traffic speed of road segments passed during the trip, km/h) & 96.16  & 7.32  & 123.00  & 71.00   \\ 
        \textbf{risk label} (Forward collision risk on target segments) & 0.04  & 0.20  & 1.00  & 0.00  \\ \hline
    \end{tabular}
    \label{tab: Statistical description}
\end{table}

\subsection{Literature collection and knowledge base construction}
\label{ss:Literature collection and knowledge base construction}
\noindent
We used the search terms ``truck risk'', ``truck safety'', ``highway'', ``urban roads'', ``forward collision'', ``truck driver'', ``fatigue'', ``distraction'' and their combinations to retrieve relevant literature. In the construction pipeline of the literature knowledge base, the LLM performs full-text screening and determines whether each paper is relevant to the research topic of truck driving risk prediction. Therefore, we downloaded all available articles from the database that could be retrieved using the aforementioned search terms. Finally, 299 research papers were downloaded in the format of PDF.\par 

\begin{figure}[htb]
\centering
\includegraphics[width=\textwidth]{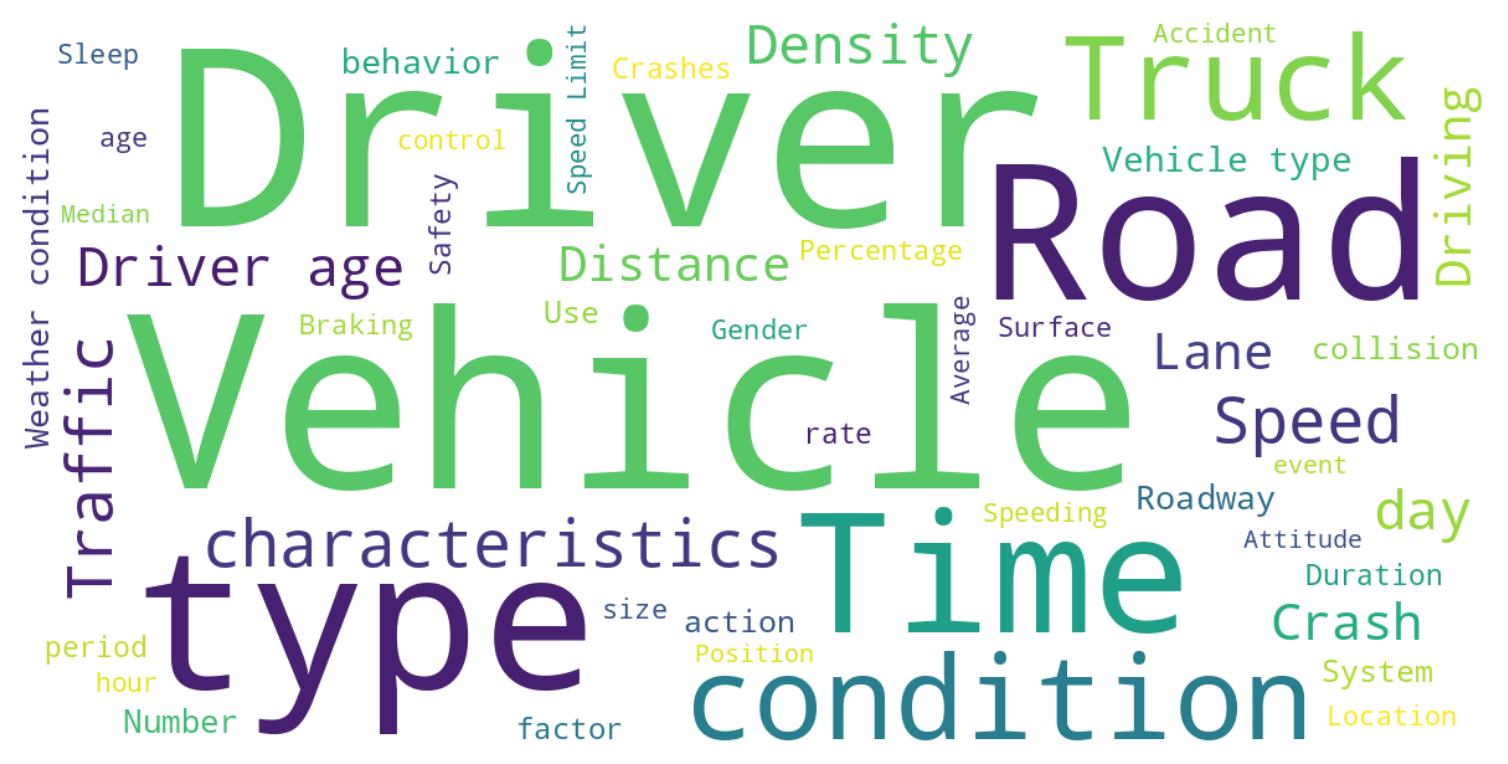}
\caption{Word cloud of the influencing factors studied in the 137 related papers.}
\label{fig:Word cloud}
\end{figure}

After downloading the papers in PDF format, we first used Marker to convert the PDF files into Markdown files. In the Markdown files, formatting elements such as sections, tables, numbered lists, and bold text are preserved, and figures from the papers were ignored. Subsequently, we used a Qwen2.5-7B-Instruct with extended context to screen the papers and judge if the paper is relevant to this study. Specifically, we used YaRN technique \citep{peng2023yarn} to extend the context length of Qwen2.5-7B-Instruct from 32k tokens to 128k tokens so that it can process the whole research papers. Using the prompt as shown in Figure \ref{fig:Prompt for literature information extraction}, we extracted LLM's answer about each paper, including if the paper is studying the potential impacting factors of truck safety, and dropped the papers that are not relevant to this specific research topic. After this process, 137 papers relevant to this study were selected. Upon review, these research papers investigated various influencing factors of truck driving risk in diverse traffic scenarios, and the word cloud of these factors is shown in Figure \ref{fig:Word cloud}. \par

For these papers, we stored the key information extracted by the LLM in JSON format, including the paper's hypotheses, data collection conditions, identified factors influencing truck driving risk, and a conclusion of the paper. As discussed in Section \ref{ss:Literature Processing Pipeline}, we jointly input the JSON-formatted summaries from the 137 papers into Qwen2.5-7B-Instruct with extended 128k-token context. Meanwhile, using the prompt template provided in Figure \ref{fig:Prompt for domain knowledge extraction}, we instructed the LLM to summarize the impact of the input variables used in this study on truck driving risk, as well as their combinations. The definitions of the variables are also included in the prompt to ensure that the LLM understands the meanings of the variables. Finally, the domain knowledge base is constructed as shown in Table \ref{tab:Constructed knowledge base} in the appendix.\par

\subsection{Model training settings}
\label{ss:Model training settings}
\noindent
We selected Qwen2.5-7B-Instruct as the base model and train it using LoRA for supervised fine-tuning. We loaded the base model with bf16 precision, with LoRA parameters configured as rank 8 and alpha value 16. The main training parameters are shown in Table \ref{tab:Training Parameters}. \par 

\begin{table}[htbp]
\centering
\caption{Training parameters for LLM fine-tuning}
\label{tab:Training Parameters}
\begin{tabular}{lc}
    \hline
    Parameter & Value \\
    \hline
    Per device train batch size & 1 \\
    Gradient accumulation steps & 8 \\
    Learning rate & 1.0e-4 \\
    Num train epochs & 3.0 \\
    LR scheduler type & cosine \\
    Warmup ratio & 0.1 \\
    \hline
\end{tabular}
\end{table}

Before training, the dataset is divided into training set and testing set using 1:1 ratio. And the LLM and the benchmark models are trained on the training set with SMOTE technique to balance the risky and non-risky samples. \par

\subsection{Benchmark models}
\label{ss:Benchmark models}
\noindent
We selected the following benchmark models to verify the performance of LLMs after fine-tuning and knowledge enhancement. Random forest \citep{breiman2001random} is a widely-used ensemble learning method for classification tasks that constructs multiple decision trees during training. XGBoost \citep{chen2016xgboost} is a gradient boosting framework that implements the gradient boosting decision tree algorithm with additional regularization terms to prevent overfitting. MLP (Multi-Layer Perceptron) \citep{hornik1989multilayer} is a widely-used neural network architecture which learns non-linear relationships in the data through backpropagation. The hyperparameters of the above benchmark models were optimized for stable predictive performance on the training set with SMOTE technique.\par

\subsection{Evaluation metrics}
\label{ss:Evaluation metrics}
\noindent
We used the overall accuracy metric to evaluate the prediction performance of the model on all samples. Considering that the prediction performance of the model on positive samples (real risk events) is more important than that of negative samples (no risk events) for the risk prediction problem, we also use precision, recall, and F1 score as evaluation metrics. The definitions of these metrics are as follows:
\begin{align}
\label{eq:accuracy}
\text{Accuracy} &= \frac{TP + TN}{TP + TN + FP + FN} \\
\label{eq:precision}
\text{Precision} &= \frac{TP}{TP + FP} \\
\label{eq:recall}
\text{Recall} &= \frac{TP}{TP + FN} \\
\label{eq:f1-score}
\text{F1-Score} &= \frac{2 \times \text{Precision} \times \text{Recall}}{\text{Precision} + \text{Recall}}
\end{align}

where TP, TN, FP, and FN represent the number of true positives, true negatives, false positives, and false negatives, respectively. In the risk prediction system, higher precision indicates a lower false alarm rate and higher recall indicates a lower miss rate, and the F1-score reflects the balance between the two.\par

\section{Result analysis}
\label{s:Result analysis}
\noindent

\subsection{Task 1: Prediction performance}
\label{ss:Prediction performance}
\noindent

In the evaluation, we sampled risky and non-risky trajectories using the sample ratio of 1:4 from the test set to suit the testing standard in the industry. Subsequently, we tested the models on the test set composed of these samples, including 44 trajectories with forward collision risk on the target segments and 176 trajectories without such risk.\par

The overall prediction performance of the LIFT LLM and the benchmark models on the test set are shown in Table \ref{tab:prediction metrics}. The LIFT LLM exceeded all benchmark models in accuracy of the truck driving risk prediction problem. Furthermore, since the LIFT LLM achieved the highest 0.95 in recall---indicating 95\% of true risky cases were correctly predicted by the model---it could be particularly useful in an onboard warning system which aims to detect as many potential risk cases as possible to enable effective safety alerts. While the precision of the LIFT LLM was below the benchmark models, the 0.76 F1-score of the LIFT LLM still exceeded all the benchmark models, demonstrating its advance in balancing precision and recall. Among the benchmark models, the MLP performed closest to the LIFT LLM---particularly in terms of recall, demonstrating the advantage of neural networks over decision trees in capturing complex feature associations. Yet, the LIFT LLM still achieved 26.7\% and 10.1\% improvements over the MLP in recall and F1-score, respectively.

\begin{table}[!ht]
    \centering
    \caption{Prediction metrics of the models}
    \begin{tabular}{lccccc}
    \hline
        Model & RF & XGBoost & MLP & \makecell[c]{Not fine-tuned LLM\\(Qwen2.5-7B)} & \makecell[c]{LIFT LLM\\(Qwen2.5-7B)}  \\ \hline
        Accuracy & 0.84 & 0.85 & 0.86 & 0.74 & \textbf{0.88}  \\ 
        Precision & \textbf{0.83} & 0.75 & 0.63 & 0.07 & 0.64  \\ 
        Recall & 0.23 & 0.34 & 0.75 & 0.02 & \textbf{0.95}  \\ 
        F1-score & 0.36 & 0.47 & 0.69 & 0.04 & \textbf{0.76} \\ \hline
    \end{tabular}
    \label{tab:prediction metrics}
\end{table}

For ablation study, we also evaluated the performance of the original Qwen2.5-7B-Instruct LLM on truck driving risk prediction problem using the same evaluation method without fine-tuning process. Without fine-tuning on the textualized dataset, the accuracy dropped to 0.74 while precision, recall and F1-score all dropped to under 0.1. This result indicates that the original LLM failed to correctly predict most high-risk truck driving trajectory samples. Since the original LLM was trained on Internet data that contains virtually no information about data distribution corresponding to the truck driving risk prediction task defined in our study, this result is reasonable. Meanwhile, this result is consistent with the unsatisfying zero-shot prediction performance using LLMs in previous studies \citep{liu2024can, liu2025aligning}. \par

\subsection{Task 2: Interpretation of trajectory risk}
\label{ss:Interpretation of trajectory risk}
\noindent

\subsubsection{Key variable analysis}
\label{ss:Key variable analysis}
\noindent
To explain the high risk trajectory samples, as shown in Figure \ref{fig:Prompt template task2}, a second prompt template for Task 2 was utilized to analyze the key variables and combination of variables that contributed to the high driving risk for each high risk trajectory sample. In the response of the LIFT LLM, at least one variable was identified as the most important influencing factors leading to the high risk in the given trajectory sample. We extracted the key influencing factors in the responses for all high risk trajectory samples, and ranked them based on the frequency of their occurrences. We conducted 10 random trials on the high-risk trajectory samples and averaged the results to mitigate the impact of randomness.\par

\begin{figure}[htb]
\centering
\includegraphics[width=\textwidth]{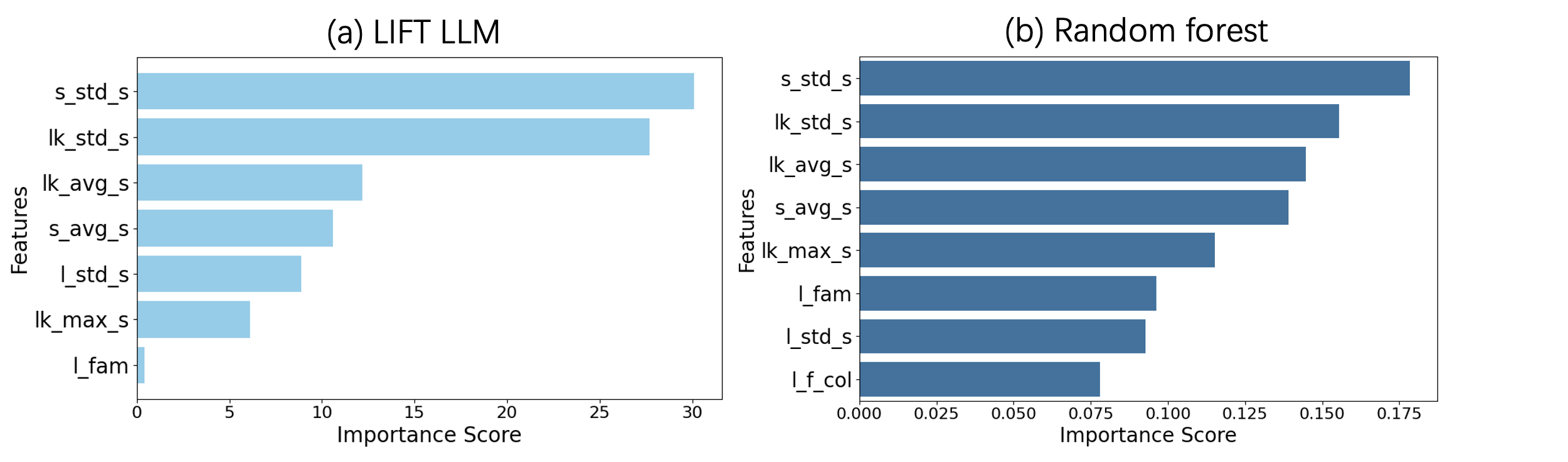}
\caption{Comparison of feature importance ranking results of LIFT framework and random forest.}
\label{fig:explanation-variables}
\end{figure}

Figure \ref{fig:explanation-variables}(a) shows the ranking result of the identified contributing factors leading to the high risk in the high-risk trajectory samples. A total of 7 features are identified as the most influencing factors among the high-risk trajectory samples, and standard deviation of driving speed during the trip (s\_std\_s) and standard deviation of traffic speed of road segments passed during the trip (lk\_std\_s) are the top two influencing factors leading to truck driving risk of forward collision in the dataset. These two variables are closely related with traffic speed volatility. Since abrupt changes in traffic speed are often associated with an increased risk of forward collisions, this result is consistent with common knowledge. \par

Furthermore, while long-term driving behavior in historical trips, short-term driving behavior and real-time traffic speed during the trip all have an impact on truck driving risk of forward collision, real-time information derived from short-term driving speed and real-time traffic speed has stronger impact---the top four influencing factors are all real-time information derived from short-term driving speed and real-time traffic speed during the trip, including standard deviation of driving speed during the trip (s\_std\_s), standard deviation of traffic speed of road segments passed during the trip (lk\_std\_s), average traffic speed of road segments passed during the trip (lk\_avg\_s) and average driving speed of the trip (s\_avg\_s). This result indicates that real-time driving risk prediction based on real-time data can more effectively identify the sources of risk, compared to traditional risk assessment methods based on driver profiling. \par

To validate the risk interpretation results, inspired by previous research on interpretable prediction methods \citep{yuan2022application, jin2023realtime}, we compare the variable importance ranking with that derived from random forest models. Figure \ref{fig:explanation-variables}(b) shows the feature importance results derived from the random forest model trained on the same dataset. The feature importance score is averaged by 10 times of random test of RF modeling. Results show that the top four important variables identified by both methods are consistent in the ranking result, validating the capability of the LIFT LLM in capturing the most important influencing factors leading to truck driving risk. \par

\subsubsection{Key combinations of impacting variables}
\label{ss:Key combinations of impacting variables}
\noindent
Using the proposed framework, unlike the feature importance derived from the RF model, the LIFT LLM can further identify combinations of variables that have a significant impact on the high risk in trajectory samples. According to previous studies, accidents are often associated with the interplay of multiple factors. Therefore, identifying the combined effects of multiple factors could be valuable for truck safety analysis.\par

\begin{figure}[h]
\centering
\includegraphics[width=.8\textwidth]{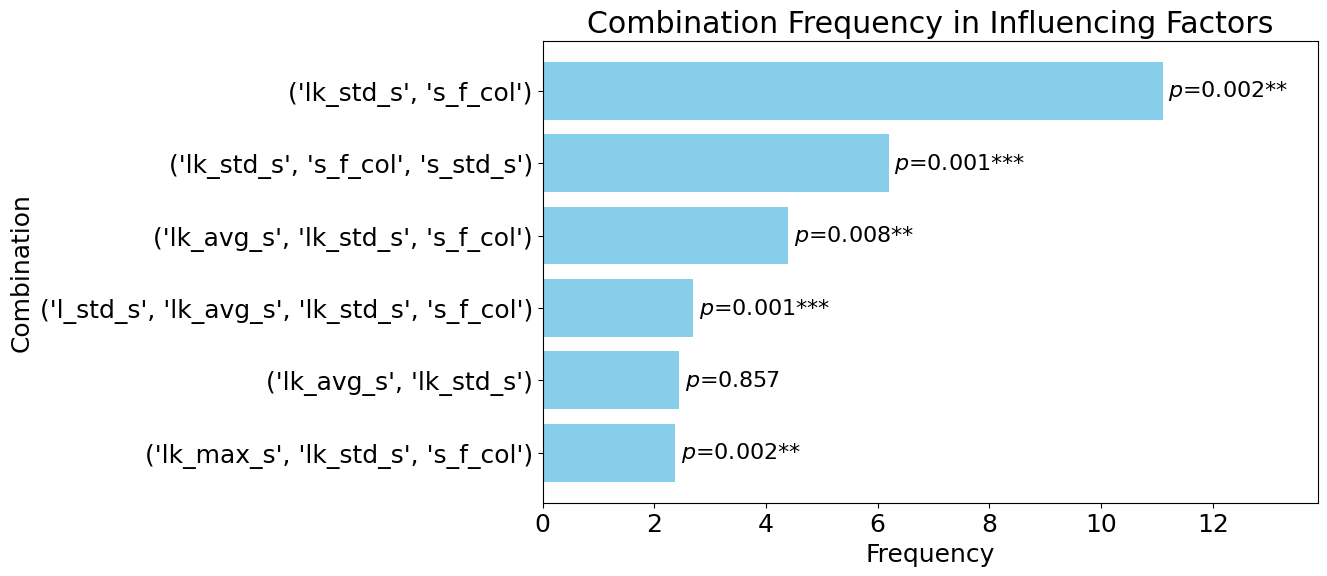}
\caption{Ranking of the most important variable combinations contributing to high-risk trajectory samples. The p-value behind each bar represents the significance of the variable combination in the PERMANOVA test between risk and non-risk samples, with *, **, and *** indicating significance levels of 0.05, 0.01, and 0.001, respectively.}
\label{fig:explanation-combinations}
\end{figure}

Figure \ref{fig:explanation-combinations} shows the most important variable combinations detected by the LIFT LLM. The ranking of variable combinations is based on the average frequency with which each variable combination was identified as a key contributing factor by the LIFT LLM across 10 random trials. While there is a vast number of possible variable combinations, only those identified more than twice on average across the trials are shown in the figure. \par

Furthermore, to validate the identified key variable combinations, we conducted a PERMANOVA test \citep{anderson2014permutational} for each combination to examine whether there are significant differences in the variable combination between risk and non-risk samples in the dataset. The p-value for the test result of each variable combination is displayed behind the corresponding bar in the figure. Results show that among the top 6 variable combinations, 5 variable combinations show significant difference between risk and non-risk samples, which validates the effectiveness of the identification results.\par

Results show that the combination of standard deviation of traffic speed of road segments passed during the trip (lk\_std\_s) and frequency of forward collision warning during the trip (s\_f\_col) is the most common influencing combination identified in the high risk trajectory samples. This result suggests that real-time traffic conditions and drivers' short-term behavioral characteristics may jointly influence driving risk. For example, when road segment speed fluctuates, the absence of short-term forward collision warnings may lead drivers to neglect maintaining a safe following distance, thereby increasing the risk of front-end collisions. Furthermore, the combination of lk\_std\_s and s\_f\_col could be a basic combination while adding more variables into the combination may constitute different scenarios which may have significant impact on driving risk.\par

For example, in the second important variable combination, the combination of standard deviation of driving speed during the trip (s\_std\_s) and the above two variables is also identified as a high-frequency combination of variables leading to high risk trajectories. This suggests that the short-term standard deviation of driving speed can further indicate the driving behavior of the truck driver, supporting better identification of truck driving risk in a potential risky combination of road traffic conditions and absence of short-term forward collision warnings. Furthermore, still on the basis of the combination of lk\_std\_s and s\_f\_col, their combination with average traffic speed of road segments passed during the trip (lk\_avg\_s) also constitutes different scenarios which may have significant impact on driving risk. This indicates that the effects of traffic speed variability and the short-term absence of forward collision warnings on driving risk may vary across different average speed conditions. And considering standard deviation of long-term driving speed (l\_std\_s) with the combination of the above three variables further constitutes new scenarios including the long-term driving style of the driver.\par

These results reveal heterogeneity in truck driving risk---samples may still exhibit variability across other factors even when certain variables are similar, leading to diverse traffic scenarios where the underlying mechanisms of risk formation may differ. The LIFT LLM effectively uncovers such heterogeneity by identifying key combinations of variables, supporting deeper investigation into these distinct risk patterns.\par

\section{Discussion on interpretable prediction with LIFT LLM}
\label{s:Discussion}
\noindent

\subsection{Contribution of the fine-tuning process and the literature knowledge base}
\label{ss:Contribution of the fine-tuning process and the literature knowledge base}
\noindent

\begin{figure}[htb]
\centering
\includegraphics[width=\textwidth]{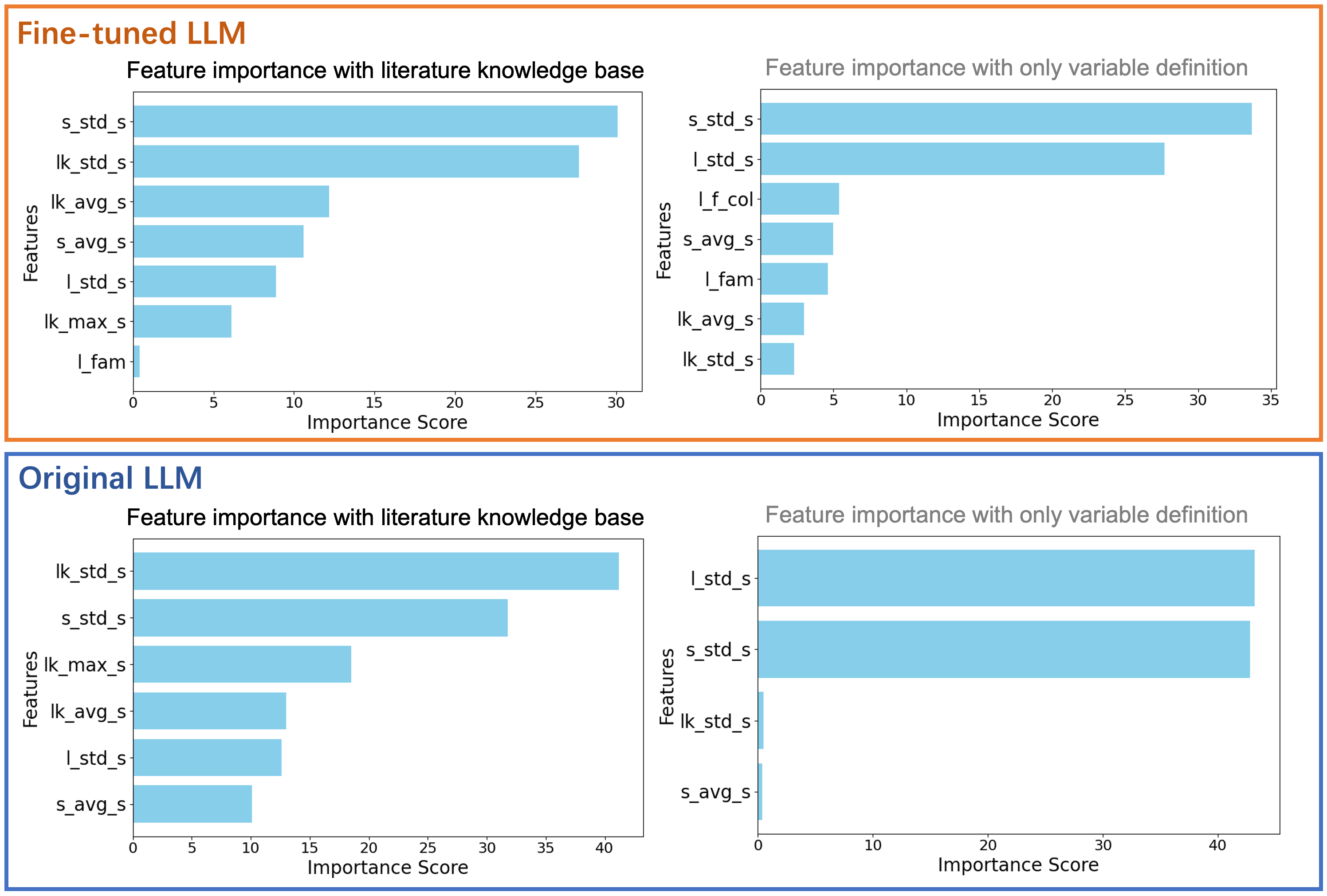}
\caption{Comparison of influencing risky factor identification results with and without literature knowledge base using the fine-tuned and not fine-tuned LLM.}
\label{fig:explanation-variables comparison}
\end{figure}

\begin{figure}[htb]
\centering
\includegraphics[width=\textwidth]{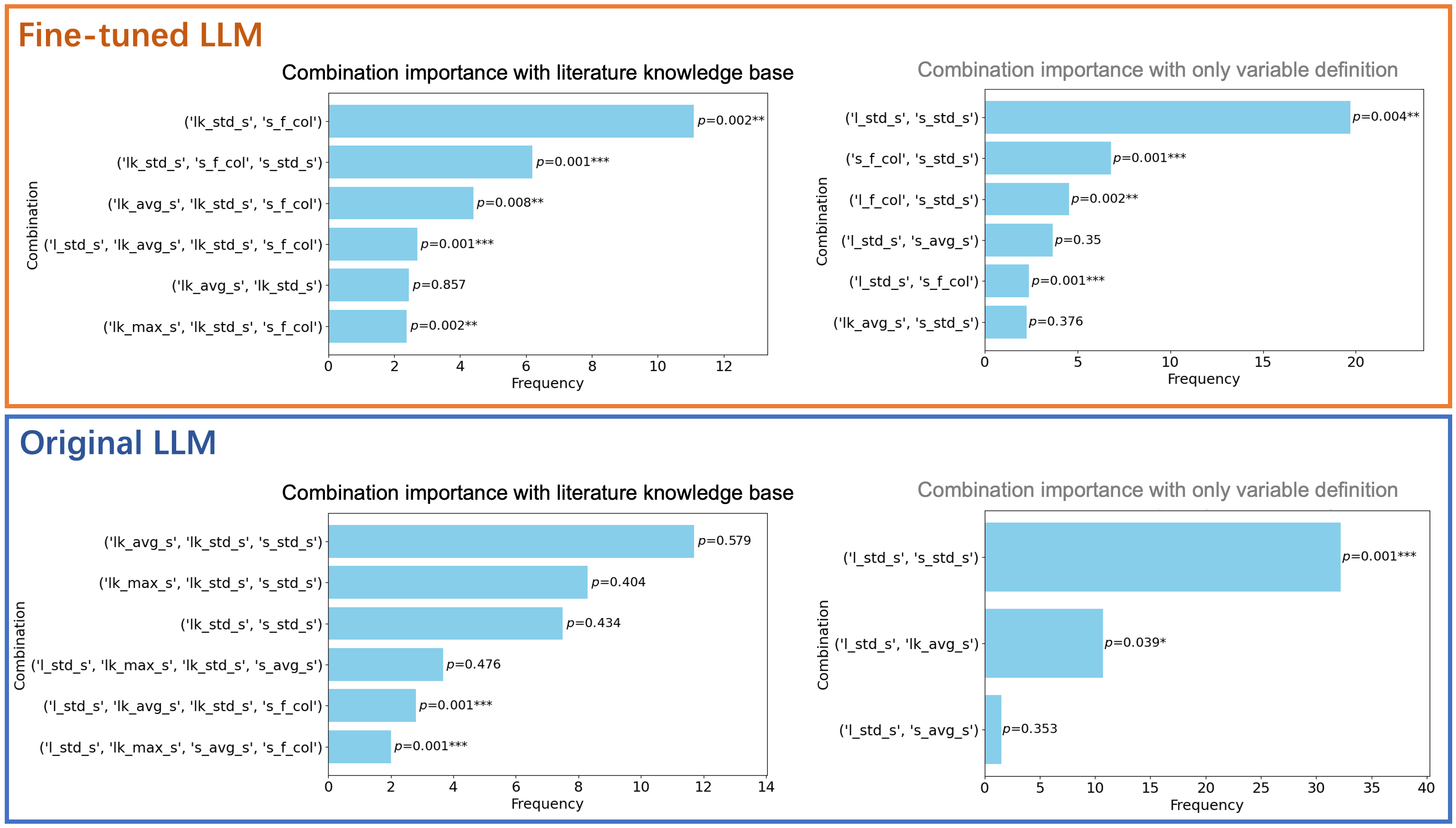}
\caption{Comparison of influencing combination of risky factors identification results with and without literature knowledge base using the fine-tuned and not fine-tuned LLM. The \textit{p}-value behind each bar represents the significance of the variable combination in the PERMANOVA test between risk and non-risk samples, with *, **, and *** indicating significance levels of 0.05, 0.01, and 0.001, respectively.}
\label{fig:explanation-combination comparison}
\end{figure}

In this section, we conduct further analysis on the contribution of the fine-tuning process and the literature knowledge base to the LIFT LLM, aiming to investigate the mechanism how these designs enhance the its ability to interpret risk causation. Specifically, we use the proposed framework to identify the most important factors and variable combinations leading to the high risk in trajectory samples under different settings, and the results are shown in Figure \ref{fig:explanation-variables comparison} and \ref{fig:explanation-combination comparison}. All results in the figures represent the averages across 10 random trials under each setting, using a temperature factor of 0.5.\par

\subsubsection{Key variable identification}
\label{sss:Key variable identification}
\noindent
 
Figure \ref{fig:explanation-variables comparison} shows the risky factor identification results with and without literature knowledge base and fine-tuning process. In the ablation study of literature knowledge base, we deleted the literature knowledge describing the impact and combination impact the the variables on truck driving risk from the prompt, but only left the variable definition in the prompt. Main observations in Figure \ref{fig:explanation-variables comparison} are as follows:

\begin{itemize}
    \item When both fine-tuning process and the literature knowledge base are removed, the original LLM only output constant explanation results (i.e., s\_std\_s and l\_std\_s) for all high risk trajectory samples in most trials.
    
    \item When the literature knowledge base is removed, the fine-tuning process slightly alters the LLM, leading it to identify a broader range of contributing factors in result interpretation. However, s\_std\_s and l\_std\_s remain the two most frequently identified variables.

    \item When the fine-tuning process is removed, the literature knowledge base effectively helps the LLM identify a broader range of contributing factors. In the resulting rankings, the top two most important factors are consistent with those identified by RF in Figure \ref{fig:explanation-variables}(b), though their order differs.
    
\end{itemize}

For the influencing factor identification task, the original LLM was unable to capture the complex variables that impacting truck driving risk based on the provided data. The literature knowledge base provides domain knowledge regarding how these variables influence truck driving risk according to previous research, effectively enhancing the LLM's identification capability. However, as the knowledge in the literature is acquired from datasets collected under conditions different from the that of the specific dataset used in this research, the knowledge from previous data may not perfectly interprets the risk results in the new dataset. Under such conditions, the fine-tuning process further adapts the LLM to the data distribution in this study, improving the accuracy of factor identification and achieving results that are largely consistent with those obtained by the RF model in Figure \ref{fig:explanation-variables}(b).\par

\subsubsection{Key variable combination identification}
\label{sss:Key variable combination identification}
\noindent

Figure \ref{fig:explanation-combination comparison} shows the key variable combination identification results of the proposed framework with and without literature knowledge base and fine-tuning process. Main observations are as follows: 

\begin{itemize}
    \item When both fine-tuning process and the literature knowledge base are removed, the original LLM identifies only 2–3 fixed combinations in most trials.

    \item Similar to the conclusions for influencing risky factor identification results in Figure \ref{fig:explanation-variables comparison}, both the literature knowledge base and the fine-tuning process enhance the LLM to identify a broader range of variable combinations.

    \item Comparing with the fine-tuning process, the literature knowledge base enables the model to identify more complex variable combinations.

    \item Comparing with the literature knowledge base, the fine-tuning process enables the model to identify more variable combinations that exhibit significant differences between risk and non-risk samples.
    
\end{itemize}

In conclusion, the comparative results in this section show that both the literature knowledge base and the fine-tuning process enhance the LLM's ability to interpret high-risk samples. The literature knowledge base provides the LLM with deeper domain understanding, while the fine-tuning process enables the LLM to better adapt to the distribution of the provided data. Furthermore, as the above result shows that the knowledge base can significantly affect the risky factor identification results, it highlights the importance of customizing the knowledge base according to industry experience and specific operational requirements in industrial applications, which would enable the identification results to better support real-world operational needs. The related discussion is provided in Section \ref{sss:Potential of customized knowledge base in truck safety management}. \par

\subsection{Robustness of result explanation}
\label{ss:Robustness of result explanation}
\noindent

This section primarily discusses the robustness of the LIFT LLM in generating causal explanations from given literature knowledge base---specifically, whether it can produce consistent results under identical inputs of literature knowledge and high-risk event datasets. And we further compare the robustness of result explanations between the LIFT LLM and the benchmark method, considering the stability in model inference and the robustness against heterogeneity in data sampling conditions.\par

\subsubsection{Stability of model inference}
\label{sss:Stability of model inference}
\noindent

\begin{figure}[htb]
\centering
\includegraphics[width=\textwidth]{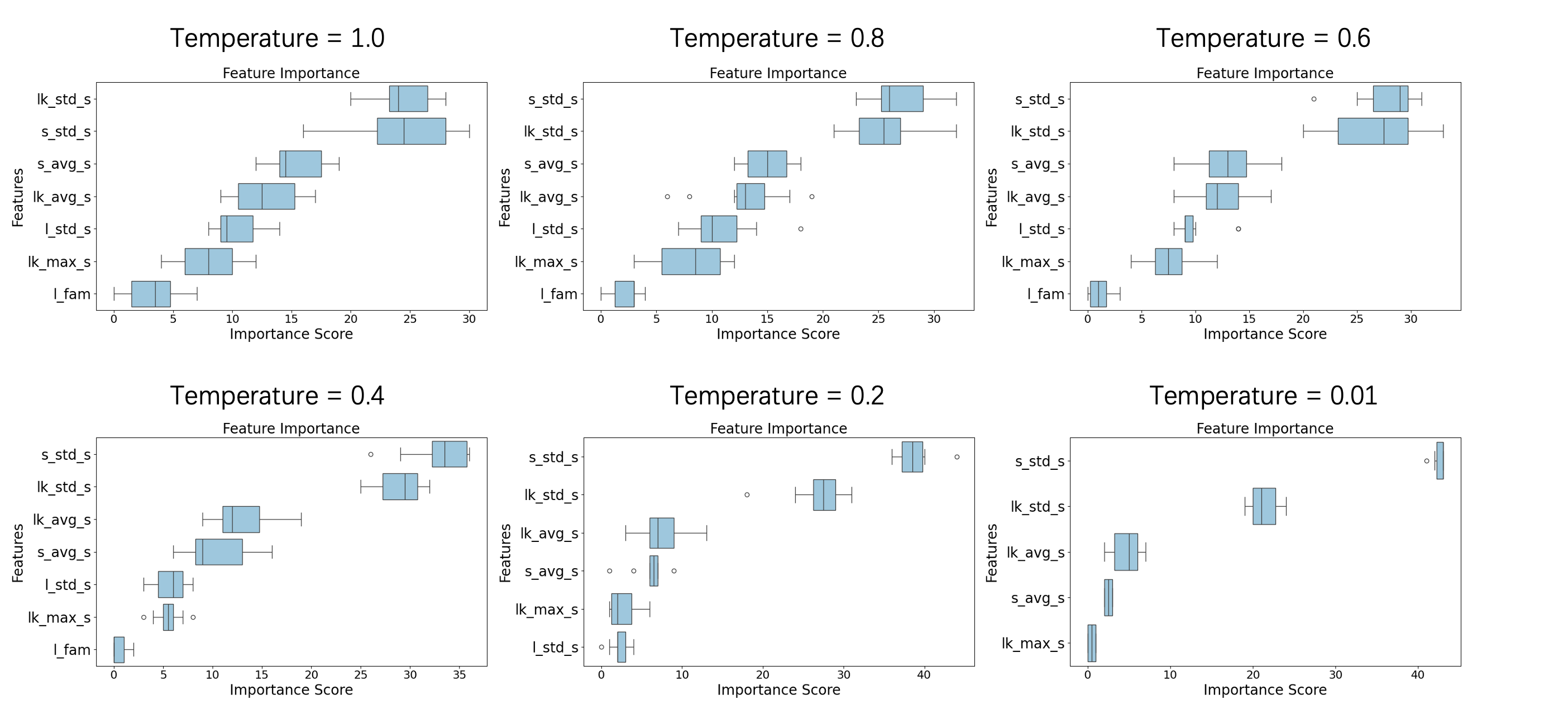}
\caption{Distribution of feature importance in 10 times of random tests using LIFT LLM with different temperature settings.}
\label{fig:different temperature}
\end{figure}

In the inference process of LLM, as shown in Eq. (\ref{eq:NTP with temperature}), the temperature parameter $\tau$ affects the randomness of the output. By setting a lower temperature parameter $\tau$, we can enhance the response stability of the LLM, ensuring that the LLM produces consistent outputs given the same input prompts. Figure \ref{fig:different temperature} presents the distribution of variable importance identification results under different temperature settings.\par

When the temperature is set to 1.0, the randomness in the output of the LIFT LLM is high, and there is considerable overlap in the importance distributions of some variables (e.g., lk\_std\_s and s\_std\_s), suggesting instability in their ranking order across trials. As the temperature is gradually reduced to 0.01, the overlap between variable distributions gradually decreases, indicating improved consistency and stability of results across multiple trials. Meanwhile, as the temperature decreases, the number of different influencing factors identified by the model also decreases. To achieve stable results containing diverse influencing factors, this study adopts a temperature setting of 0.5 for all interpretation tasks, and use the average results for analysis. \par

\begin{figure}[htb]
\centering
\includegraphics[width=.6\textwidth]{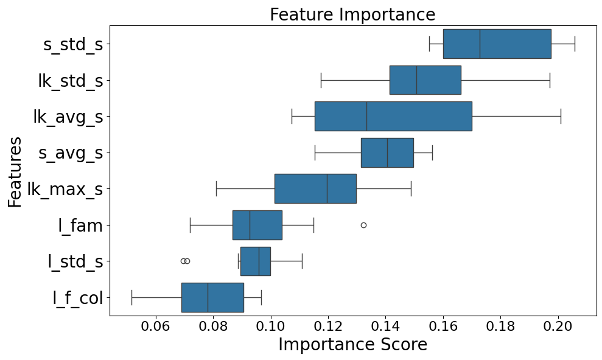}
\caption{Distribution of feature importance using random forest in 10 times of random tests.}
\label{fig:RF inherent stability on whole data}
\end{figure}

For comparison, Figure \ref{fig:RF inherent stability on whole data} shows the distribution of variable importance obtained by a random forest model over 10 random trials on the same dataset. Compared to the random forest model, the LIFT LLM with a temperature under 0.6 produces more consistent variable importance rankings across different trials.\par

\subsubsection{Robustness against heterogeneity in data sampling conditions}
\label{sss:Robustness against randomness in data sampling conditions}
\noindent

It should be noted that the LIFT LLM derives its analytical results using only the positive samples from the dataset. However, traditional machine learning methods such as random forest require both positive and negative samples from the dataset for model training. As a result, the performance of such ML methods is inherently sensitive to the data distribution of the training data in different traffic scenarios. \par

\begin{figure}[htb]
\centering
\includegraphics[width=.6\textwidth]{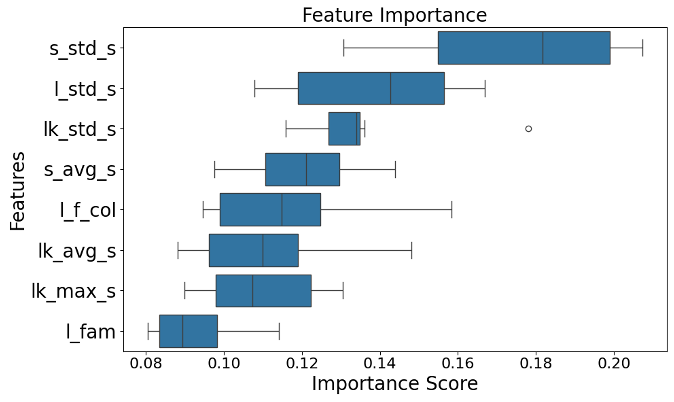}
\caption{Distribution of feature importance in 10 times of random dataset tests with different data sampling results. The feature importance result in each of the random dataset test is the averaged result in 10 times of random test using the same sampled dataset.}
\label{fig:RF data stability}
\end{figure}

For instance, Figure \ref{fig:RF data stability} illustrates the distribution of variable importance obtained by the random forest model in 10 experiments with different data sampling conditions. In each experiment, negative samples are randomly selected from the test set in a 1:1 ratio relative to the number of positive samples. A random forest model is then trained on each of the resulting sampled datasets, and the average variable importance is computed over 10 model runs. Finally, the distributions of the average variable importance across the 10 independent sampling iterations are visualized as boxplots in Figure \ref{fig:RF data stability}. As shown in the figure, even though the variable importance is averaged over 10 model runs for each data sampling, there remains considerable variability in the variable importance across different sampling iterations.\par

In existing practices, when data is sufficiently rich or the data distribution is stable, the interpretation results such as variable importance provided by traditional machine learning methods are widely accepted. However, when data collection conditions vary significantly---for instance, across different road classes, weather conditions, or even extreme weather scenarios---the analysis results from traditional methods may lack consistency. In such cases, just as illustrated in this study, an LIFT LLM-based approach grounded in a literature knowledge base may offer more stable and robust interpretations.\par

\subsection{Potential of LIFT LLM}
\label{ss:Potential of LIFT framework}
\noindent

\subsubsection{Potential of customized knowledge base in truck safety management}
\label{sss:Potential of customized knowledge base in truck safety management}
\noindent

In traffic safety domain, a substantial body of academic research has analyzed truck driving risks across diverse traffic scenarios. In this study, we have illustrated that the LIFT LLM effectively leverages such domain knowledge from the literature to enable real-time causal factor identification in high-risk scenarios. Compared with data-dependent ML methods, factor identification driven by the LIFT LLM offers greater stability in results regardless of data sampling conditions. Moreover, the model can identify key variables and variable combinations for each high-risk situation in a real-time manner, thereby exhibits its potential in assisting safety intervention teams in providing precise intervention to truck drivers.\par

Furthermore, beyond idealized academic settings, industry practitioners have accumulated extensive operational experience in truck safety management. This practical expertise can be integrated into customized knowledge bases to improve the relevance and utility of risk prediction models in real-world applications. For example, for truck fleets in mountainous areas, the knowledge base can be customized to contain more details about long downhill risks and sharp turning risks; for fleets that frequently operate at night, the knowledge base can include more information related to fatigue driving. Using the proposed interpretable prediction framework with LIFT LLMs, logistics companies can tailor knowledge bases and fine-tune models according to specific operational needs, while systematically evaluating both predictive performance and model interpretability to ensure alignment with safety management requirements. Consequently, the proposed framework in this work enables logistics companies to fully exploit existing expert knowledge and real-world data, reduce the workload of intervention teams, and enhance the overall quality of risk mitigation services.\par

\subsubsection{Potential of LIFT LLM in risky scenario discovery}
\label{sss: Potential of LIFT framework in risky scenario discovery}
\noindent

As safety standards in the logistics industry continue to rise, logistics companies need to conduct fine-grained risk management for different traffic scenarios in freight transportation. In current practice, potential risk scenarios are often identified based on expert experience, such as rainy weather or traffic speed fluctuations. However, existing studies have revealed substantial heterogeneity in traffic risk scenarios \citep{behnood2019timeofday, yuan2021risk, hu2025influencing}. For instance, even under the same traffic speed fluctuation condition, drivers with different driving styles may face distinct levels of risk. With the increasing availability of multi-source data, identifying risk scenarios may require the combination of dozens of different variables, which is difficult to accomplish through expert experience alone.\par

In this study, we have shown the potential of the LIFT LLM in discovering new risky scenarios among the dataset. Furthermore, by identifying more complex variable combinations, the LIFT LLM revealed heterogeneity in the typical risky scenario depicted by the variable combination of standard deviation of traffic speed of road segments passed during the trip and frequency of forward collision warning during the trip. And we used PERMANOVA test to validate the significance of the difference of the complex variable combinations between risk and non-risky samples, which suggests that these complex combinations might be potential risky scenarios for truck safety. This study has shown the potential of the LIFT LLM as a new research direction of discovering new risky scenarios, which may remarkably accelerate scientific discovery and safety improvements in the field of truck safety. \par

\section{Conclusions}
\label{s:Conclusions}
\noindent
In this research, we proposed a novel interpretable prediction framework with literature-informed fine-tuned (LIFT) LLMs. The framework is centered around an \textit{Inference Core}, where the LLM is fine-tuned on a real-world truck driving risk dataset to achieve accurate prediction and reliable interpretation of truck driving risk. In addition, the framework integrates a \textit{Literature Processing Pipeline} that filters and summarizes domain-specific literature into a literature knowledge base, and a \textit{Result Evaluator} which evaluates the prediction performance as well as the interpretability of the LIFT LLM.\par

We adopted an open-source LLM, Qwen2.5-7B-Instruct, in the framework and fine-tuned it using a real-world truck driving risk dataset. Evaluation results showed that the fine-tuned LLM can precisely predict truck driving risk according to the real-time driving data, outperforming benchmark models. Based on 299 research papers in the truck driving safety domain, the \textit{Literature Processing Pipeline} constructed a domain literature knowledge base containing knowledge about the impact and combination impact of the risk variables on truck driving risk in the dataset. With the literature knowledge base, we prompted the LIFT LLM to identify the most important variables and combination of variables leading to high risk in the trajectory samples. Results showed that the LIFT LLM produced variable importance rankings highly similar to those derived from the widely-adopted random forest model. Furthermore, the LIFT LLM identified complex variable combinations that showed significant differences in PERMANOVA tests between risk and non-risk samples, indicating potential heterogeneous truck risk scenarios.  \par

We further revealed how the integration of the literature knowledge base and the fine-tuning process enhanced the LLM’s ability to interpret truck driving risk. The knowledge base provides foundational knowledge about variable associations drawn from the literature, enriching the LLM’s understanding beyond its original pretraining data and strengthening the connections between various factors and truck driving risk. Meanwhile, the fine-tuning process improves the LLM’s capability to identify and interpret risk factors based on the data distribution of a specific dataset, enabling it to discover new knowledge from new datasets. We discussed the robustness of explanation capability of the LIFT from two perspectives: the stability of model inference and robustness against heterogeneity in data sampling conditions, and illustrated the advantage of the LIFT LLM against traditional models. And the potential of the LIFT LLM in truck safety management and risky scenario discovery is illustrated by concrete examples.\par

Through the application on truck driving risk prediction, this study illustrates the potential of LIFT LLMs in data-driven knowledge discovery, providing the following insights.

\begin{itemize}
    \item A general-purpose LLM fine-tuned on domain-specific data can achieve state-of-the-art performance in the specialized task of truck driving risk prediction.

    \item A domain literature knowledge base is valuable for a general-purpose LLM to understand domain problems and provide explanations on prediction results.

    \item Combining a literature knowledge base with fine-tuning on real-world datasets enables LLMs to discover new knowledge based on prior understanding and the underlying data distribution, such as discovering heterogeneity in risky traffic scenarios.
\end{itemize}

In addition to risk scenario discovery in traffic safety, the LIFT LLM can also be applied to knowledge discovery in other domains. Yet, the issue of hallucination remains to be addressed. In this study, we mitigate random generation by explicitly providing a literature knowledge base in the prompt and using a low temperature setting, guiding the LIFT LLM to produce responses that are more consistent with the provided domain knowledge and the data distribution. Alternative approaches, such as pretraining LLMs on domain-specific literature, could be further explored. Yet, as such pretraining demands refined training techniques and substantial computational resources, it is left for future work. Furthermore, future studies may also conduct fine-grained quantitative analyses of potential hallucinations in LLMs and explore better designs for the literature knowledge bases.\par

\section*{CRediT authorship contribution statement}
\noindent
\textbf{Xiao Hu}: Conceptualization, Methodology, Software, Writing - original draft. \textbf{Yuansheng Lian}: Software, Writing - review \& editing. \textbf{Ke Zhang}: Conceptualization, Writing - review \& editing. \textbf{Yunxuan Li}: Conceptualization, Writing - review \& editing. \textbf{Yuelong Su}: Conceptualization, Writing - review \& editing. \textbf{Meng Li}: Conceptualization, Writing - review \& editing.

\section*{Declaration of Competing Interest}
\noindent
The authors declare that they have no known competing financial interests or personal relationships that could have appeared to influence the work reported in this paper.\par

\section*{Acknowledgments}
\noindent
This research is supported by National Natural Science Foundation of China No.52325209 and Independent Research Project of the State Key Laboratory of Intelligent Green Vehicle and Mobility, Tsinghua University (No. ZZ-GG-20250406). We gratefully acknowledge SFMAP Technology and Amap for providing essential data support for this research.\par

\bibliographystyle{elsarticle-harv}
\bibliography{main}

\newpage
\appendix
\section{The constructed literature knowledge base}
\label{apdx: The constructed literature knowledge base}
\noindent

\setcounter{figure}{0}
\setcounter{table}{0}
\renewcommand{\thesection}{\Alph{section}}
\renewcommand{\thefigure}{\thesection-\arabic{figure}}
\renewcommand{\thetable}{\thesection-\arabic{table}}

\begin{table}[h]
\tiny
\centering
\caption{Constructed knowledge base. For ease of reading, the constructed knowledge base is presented in tabular form. In actual model inference, the knowledge base is input in JSON format.}
{
\renewcommand{\arraystretch}{1.2}
\linespread{1.2}\selectfont
\begin{tabular}{|p{1.5cm}|p{1.5cm}|p{12cm}|}
\hline
\textbf{Influencing factor} & \textbf{Type} & \textbf{Content} \\
\hline
\multirow{3}{*}{\centering s\_f\_col} 
    & definition & frequency of forward collision warning during the trip \\ \cline{2-3}
    & impact & A higher frequency of forward collision warnings is associated with a higher risk of actual forward collisions, suggesting that attention to these warnings can help reduce collision likelihood. \\ \cline{2-3}
    & combination impact & When combined with other factors such as l\_fam and lk\_avg\_s, a high s\_f\_col can indicate a higher risk of forward collisions in previously unpassed road segments with average traffic speeds, potentially leading to increased alertness and safety measures. \\ \hline

\multirow{3}{*}{\centering l\_std\_s} 
    & definition & standard deviation of longterm driving speed \\ \cline{2-3}
    & impact & A higher standard deviation of driving speed suggests varying driving conditions and can increase the likelihood of forward collisions due to inconsistent speed control. \\ \cline{2-3}
    & combination impact & When combined with s\_avg\_s and lk\_std\_s, high l\_std\_s can indicate situations where the truck's speed fluctuates significantly compared to the average speed, increasing the risk of forward collisions. \\ \hline

\multirow{3}{*}{\centering lk\_max\_s} 
    & definition & maximum traffic speed of road segments passed during the trip \\ \cline{2-3}
    & impact & Higher maximum traffic speeds may require more precise vehicle control, increasing the likelihood of forward collisions due to potential speed differences. \\ \cline{2-3}
    & combination impact & Higher lk\_max\_s in combination with l\_f\_col can indicate that this road segment is historically risky, and the current trip may be experiencing similar conditions, thus increasing the probability of forward collisions. \\ \hline

\multirow{3}{*}{\centering lk\_avg\_s} 
    & definition & average traffic speed of road segments passed during the trip \\ \cline{2-3}
    & impact & Average traffic speed impacts collision likelihood; higher speeds generally increase the risk of forward collisions. \\ \cline{2-3}
    & combination impact & Higher lk\_avg\_s combined with lk\_max\_s and s\_f\_col may increase the risk of forward collisions, as it suggests high average speeds that require more attention to forward collision warnings and road conditions. \\ \hline

\multirow{3}{*}{\centering lk\_std\_s} 
    & definition & standard deviation of traffic speed of road segments passed during the trip \\ \cline{2-3}
    & impact & A higher standard deviation of traffic speed indicates traffic variability, which can increase the likelihood of forward collisions due to unpredictable speed changes. \\ \cline{2-3}
    & combination impact & When combined with l\_std\_s and s\_avg\_s, higher lk\_std\_s can indicate unpredictable driving conditions, increasing the probability of forward collisions if drivers do not adjust to variable traffic speeds. \\ \hline

\multirow{3}{*}{\centering s\_avg\_s} 
    & definition & average driving speed of the trip \\ \cline{2-3}
    & impact & Lower average driving speeds are associated with reduced risk of forward collisions. Higher speeds may increase the risk due to the need for more precision in vehicle control. \\ \cline{2-3}
    & combination impact & Combined with l\_fam, a higher s\_avg\_s may indicate the truck is on a road segment frequently passed by other vehicles, suggesting that the higher speed is a known risk factor for this area. \\ \hline

\multirow{3}{*}{\centering s\_lane\_d} 
    & definition & frequency of lane departure warning during the trip \\ \cline{2-3}
    & impact & A higher frequency of lane departure warnings can indicate driver inattention or vehicle instability, both of which can increase the likelihood of forward collisions. \\ \cline{2-3}
    & combination impact & Combinations with l\_std\_s and s\_f\_col suggest that trucks experiencing lane departure warnings are also prone to higher driving speed variability and more frequent forward collision warnings, indicating a higher likelihood of forward collisions. \\ \hline

\multirow{3}{*}{\centering l\_fam} 
    & definition & ratio of historical trips that passed the target road segment \\ \cline{2-3}
    & impact & The ratio of past trips passing the road segment has no direct impact on the current trip's likelihood of a forward collision but provides context for road segment familiarity. \\ \cline{2-3}
    & combination impact & When combined with s\_f\_col and lk\_std\_s, a high l\_fam ratio may indicate a known risk in the road segment, but the presence of s\_f\_col and lk\_std\_s higher than normal for this segment can still increase the probability of forward collisions. \\ \hline

\multirow{3}{*}{\centering s\_std\_s} 
    & definition & standard deviation of driving speed during the trip \\ \cline{2-3}
    & impact & A higher standard deviation of driving speed indicates inconsistent driving, which can increase the likelihood of forward collisions due to less stable driving behavior. \\ \cline{2-3}
    & combination impact & Combined with l\_std\_s and lk\_std\_s, a high s\_std\_s suggests highly variable driving speeds and conditions, increasing the probability of forward collisions. \\ \hline
\end{tabular}
}
\label{tab:Constructed knowledge base}
\end{table}


\end{document}